\documentclass[lettersize,journal]{IEEEtran}
\usepackage{amsthm}
\usepackage{amsmath,amsfonts}
\usepackage{amssymb}
\usepackage{algorithmic}
\usepackage{algorithm}
\usepackage{array}
\usepackage[caption=false,font=scriptsize,labelfont=sf,textfont=sf]{subfig}
\usepackage{textcomp}
\usepackage{stfloats}
\usepackage{url}
\usepackage{verbatim}
\usepackage{graphicx}
\usepackage{color}
\usepackage{cite}
\usepackage{hyperref}
\usepackage{dirtytalk}
\graphicspath{{figure/}}
\newbox{\bigpicturebox}
\usepackage{multirow}

\input{bold_letter.sty}

\usepackage{bm}

\newsavebox{\measurebox}

\makeatletter
\newcommand*\titleheader[1]{\gdef\@titleheader{#1}}
\AtBeginDocument{%
  \let\st@red@title\@title
  \def\@title{%
    \bgroup\normalfont\small\raggedright\@titleheader\par\egroup
    \vskip1.5em\st@red@title}
}
\makeatother

\title{Do You Need a Hand? -- a Bimanual Robotic Dressing Assistance Scheme}
\titleheader{This paper is accepted and will appear in the IEEE Transactions on Robotics. \textcopyright 2024 IEEE. Personal use of this material is permitted.  Permission from IEEE must be obtained for all other uses, in any current or future media, including reprinting/republishing this material for advertising or promotional purposes, creating new collective works, for resale or redistribution to servers or lists, or reuse of any copyrighted component of this work in other works.}
\author{Jihong Zhu$^{1,3}$, Michael Gienger$^{2}$, Giovanni Franzese$^{3}$, and Jens Kober$^{3}$
\thanks{$^{1} $J.~Zhu is with the School of Physics, Engineering, and Technology, and Institute for Safe Autonomy, the University of York, the UK. 
        {\tt\footnotesize jihong.zhu@york.ac.uk}}
\thanks{$^{2} $M.~Gienger is with Honda Research Institute  Europe, Germany
        {\tt\footnotesize michael.gienger@honda-ri.de}}
\thanks{$^{3}$J.~Zhu,~G.~Franzese,~and~J.~Kober are with Cognitive Robotics, 3mE, Delft University of Technology, Netherlands
        {\tt\footnotesize \{J.Zhu-3,G.Franzese,J.Kober\}@tudelft.nl}}}%

\begin{document}

\maketitle

\begin{abstract}
    Developing physically assistive robots capable of dressing assistance has the potential to significantly improve the lives of the elderly and disabled population. However, most robotics dressing strategies considered a single robot only, which greatly limited the performance of the dressing assistance. In fact, healthcare professionals perform the task bimanually. Inspired by them, we propose a bimanual cooperative scheme for robotic dressing assistance. In the scheme, an interactive robot joins hands with the human thus supporting/guiding the human in the dressing process, while the dressing robot performs the dressing task. We identify a key feature: elbow angle that affects the dressing action and propose an optimal strategy for the interactive robot using the feature. A dressing coordinate based on the posture of the arm is defined to better encode the dressing policy. We validate the interactive dressing scheme with extensive experiments and also an ablation study. The experiment video is available on \url{https://sites.google.com/view/bimanualassitdressing/home}
\end{abstract}

      
\section{Introduction}\label{sec:intro}
The shortage of caregivers is a pressing challenge worldwide due to a decrease in birth rates combined with an increase in life expectancy: according to the WHO, by 2030, 1 in 6 people in the world will be aged 60 years or over. By 2050, the world’s population of people aged 60 years and older will double \cite{Ageing_WHO}. Assistive robots capable of physically aiding humans are promising to tackle this challenge. Among various tasks that the caregiver performs on a daily basis, the dressing was reported to be the greatest burden while the least automated \cite{dudgeon2008managing}. In this paper, we tackle the problem of dressing a human arm with a bimanual robot setup. 




Robot-assisted dressing is challenging as it involves direct physical interaction with flexible clothes and humans.
More specifically, it differs from a conventional human-robot interaction task such as co-manipulation or co-carrying \cite{agravante2019human, tarbouriech2019admittance,van2020predicting} where the object allows immediate force propagation. In the dressing assistance, the object i.e. the cloth, is soft and deformable, thus, the interaction force is hard to obtain. Due to the absence of direct force feedback and heavy occlusion during dressing, arm movement estimation during dressing is also difficult.

One common assumption to simplify the task is to consider static arm posture during dressing. Then the main challenge becomes adapting the dressing policy to different static postures. Learning from demonstrations (LfD) is often used in obtaining a generalized policy \cite{pignat2017learning, hoyos2016incremental, zhu2022learning}. 

However, a static arm posture is a very strong assumption. To make the proposed scheme work in practice, we need to consider dynamic poses during dressing. Since the task largely depends on postures, arm movement estimation is crucial. 

Computer vision is a popular way to track arm postures. It is also used in the early work in assistive dressing \cite{gao2015user} for tracking upper limb movement. However, the inevitable occlusions during dressing render most of the existing vision-based human posture tracking algorithms fail \cite{zhang2019probabilistic, chance2018elbows}. Additional sensory modules, such as proximity sensors may be employed for posture estimation \cite{erickson2018tracking}.

\begin{figure}[t]
    \centering
    \includegraphics[width=0.4\textwidth]{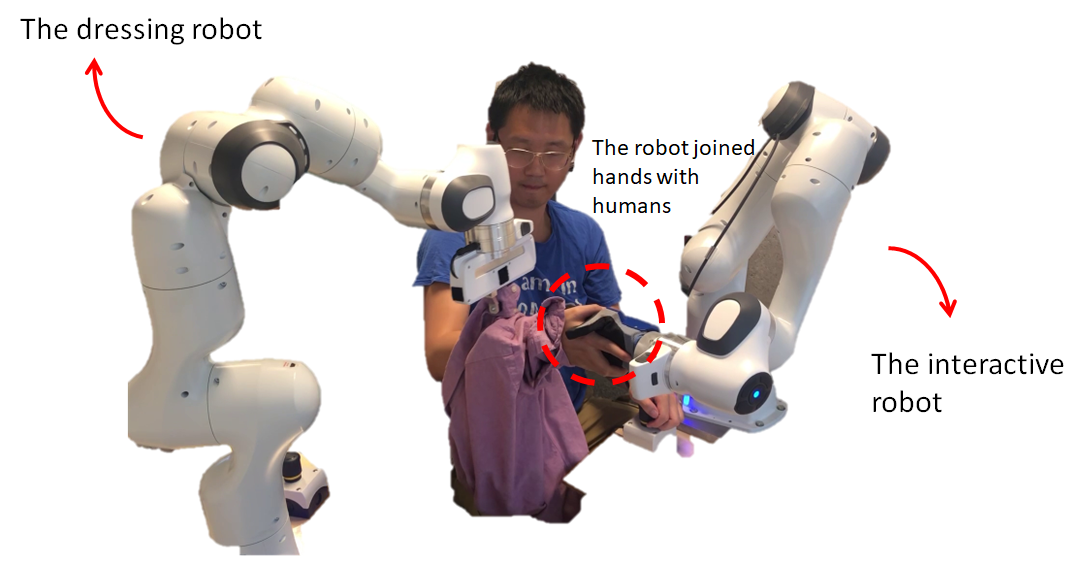}
    \caption{The bimanual robot setup for the cooperative dressing. We employ two Franka Emika robots. The dressing robot grips the cloth and executes the dressing policy. The interactive robot is equipped with a SoftHand from qbrobotics that provides human-like hand-holding. It guides the human in the dressing process.}
    \label{fig:dressing_setup_overview}
\end{figure}

To overcome the difficulties faced with conventional assistive dressing strategies, we draw inspiration from human-to-human dressing assistance from healthcare professionals\footnote{\href{https://youtu.be/Wn1OEfQt0ow?t=93}{https://youtu.be/Wn1OEfQt0ow?t=93}}. Our proposed framework (hardware setup shown in Fig. \ref{fig:dressing_setup_overview}) resembles the method adapted by health professionals. It requires two robots to perform the dressing assistance task. An interactive robot $\mathcal{I}$ that reaches out for holding hands with the human, then supports/guides the human to facilitate the dressing process. The dressing robot $\mathcal{D}$ performs the main task, dressing. Hand holding not only makes the dressing task easier to execute by supporting/guiding the human arm movement but also results in the tracking of the arm posture with proprioceptive sensors only. 

The main contribution of the paper is the proposal of a bimanual dressing assistance framework that is inspired by caregivers doing the dressing assistance task. In the framework, instead of considering interaction forces during dressing which is often hard to obtain in practice \cite{wang2022visual}, we offer a novel perspective by analysing dressing geometrically. Benefiting from this new insight, we identify a key feature: elbow angle in the human posture that affects dressing policy. The feature is then used for designing an optimal stretch policy for the interactive robot. For the dressing, in contrast to encoding the movement in Cartesian space, we define a task coordinate system for flexible policy encoding from demonstrations. 

The paper is organized as follows: In Sect.~\ref{sec:related_works} we survey the related work in three subcategories, namely: robotic dressing assistance, bimanual manipulation, and arm poster estimation. Sect.~\ref{sec:framework} provides an overview of the overall framework and the underlying assumptions. Sect.~\ref{sec:interactive_robot} introduces the feature that affects the dressing strategy and then designs an optimal strategy for the interactive robot $\mathcal{I}$ based on the feature. With the information on the hand position given by the $\mathcal{I}$, we can solve the arm posture with proprioceptive sensors only. Later in Sect.~\ref{sec:dressing_robot}, we propose a dressing coordinate dependent on arm postures, then learn the dressing policy from expert demonstrations in the dressing coordinate. The overall framework is validated with robotic experiments in Sect.~\ref{sec:experiments}. Finally in Sect.~\ref{sec:conclusion}, we conclude.
\section{Related Work}\label{sec:related_works}
\subsection{Robotic Dressing Assistance}
Robotic dressing assistance is a recurrent topic that broadly covers dressing tops (t-shirt, jacket, etc.) \cite{tamei2011reinforcement,joshi2019framework}, trousers \cite{yamazaki2016bottom} and even shoes \cite{canal2018joining,jevtic2018personalized}. Since in this paper, our focus is dressing the upper body, we review recent advances, particularly in this area.

Dressing assistance is a skill mastered by humans. Therefore, LfD-based approaches are popular in obtaining motor skills in dressing. Pignat and Calinon learned a hidden semi-Markov model \cite{pignat2017learning} from human demonstrations to encode the dressing skills. The authors of \cite{hoyos2016incremental} implemented an incremental learning of Task-Parameterized Gaussian Mixture Models (TP-GMM) that allows generalization of the dressing task. More recently, \cite{zhu2022learning} proposed a method to reduce the number of demonstrations needed for training the TP-GMM to encode a dressing policy in different static postures. 

Recent research also explores the interaction between a robot and a human in the dressing process with reinforcement learning in a simulation environment \cite{kapusta2019personalized,clegg2020learning}. The authors of \cite{yu2017haptic} learned an outcome classifier for dressing tasks with simulated haptic data. Instead of relying on simulations, data collected in the real dressing scenarios suggests that the interaction force between the cloth and the human allows for predicting dressing outcomes with high accuracy \cite{kapusta2016data}. Later, Erickson et al.\ \cite{erickson2017does} conducted an investigation with physics-based simulation suggesting that the interaction force can be recovered with only end-effector measurements.

Further research in this area is concerned with cloth state modeling \cite{twardon2016active}, safe motion planning\cite{li2021provably,li2022set}, grasping point selection \cite{zhang2020learning} and novel gripper design for dressing assistance \cite{dragusanu2022dressgripper}. 

In the above-mentioned research, the dressing tasks were mostly conducted with a single manipulator. The only papers considering bimanual dressing are \cite{clegg2020learning, tamei2011reinforcement,joshi2019framework}. However, they consider the bimanual in a completely different setup where each robot is dressing one human arm respectively. Therefore, they adopt the \textit{one-robot-to-one-arm} setup just like the rest of the papers. In our proposal, we have instead a \textit{two-robot-to-one-arm} setup, where a robot guides humans to facilitate the dressing task of the other robot. A very recent research \cite{zhang2022learning} considers also a \textit{two-robot-to-one-arm} setup while dressing the hospital gown with a mannequin lying on the bed. The task of \cite{zhang2022learning} is in a completely different setting, thus the employed framework is radically different from ours. 

\subsection{Bimanual Manipulation}
In the proposed framework, we adopt implicit master-slave coordination between the interactive and the dressing robots. The dressing policy is dependent on the human arm posture which is controlled by the interactive robot from the applied guidance force on the human hand. Such task setup is referred to as asymmetrical bimanual manipulation in the bimanual taxonomy literature \cite{krebs2022bimanual}.

The asymmetrical bi-manual task that was investigated in the previous work includes: vegetable peeling \cite{ureche2018constraints}, cloth manipulation \cite{colome2018dimensionality}, assembly \cite{suarez2018can} and cooking \cite{liu2022robot}. These tasks usually do not require interaction with a human as in the dressing assistance.

More relevant bimanual assistive tasks: Xu et al.\ designed an autonomous wheelchair with two robots to assist humans in daily tasks \cite{xu2010enhanced}. Connan et al.\ employed a humanoid robot to perform tasks such as removing the lid, unscrewing a bottle, and pouring water \cite{connan2021learning}. Edsinger and Kemp discussed three key themes in bimanual manipulation design in assistive tasks: cooperative manipulation, tasks relevant features, let the body do the thinking and analysed them in different assistive scenarios \cite{edsinger2007two}. These key themes remain very relevant to us in designing the framework for this paper.

To date, robotics dressing assistance (specifically dressing clothes) has rarely been considered a bimanual task that requires both robots. However, as we argued in Sect.~\ref{sec:intro}, the additional robot can provide support and guidance to the arm being dressed, which makes the dressing easier and more comfortable. As the scheme is employed by caregiving professionals, humans might feel more contented dressing in such a composition. By adopting the bimanual setup in dressing, our work is the first of its kind and represents a paradigm shift in thinking of the dressing assistance task.
\subsection{Arm Posture Estimation}
The dressing policy is highly dependent on human arm posture. As the guidance force at hand will change the human arm posture in our framework, a robust arm posture tracking/estimation is required for dressing policy generation.

Due to heavy occlusion during dressing, the task is particularly challenging for popular vision-based deep learning posture tracking algorithm \cite{toshev2014deeppose,cao2017realtime,guler2018densepose,iqbal2018dual}. 

Human modeling is employed for tracking the human posture under occlusions. A preliminary study in dressing assistance considers human modeling combined with visual tracking in personalized dressing: the authors of \cite{gao2015user} adopted a mixture of Gaussians for modeling, and later a Gaussian process latent variable model (GP-LVM) was employed for posture estimation in \cite{zhang2019probabilistic}. Recently, the author of \cite{chance2018elbows} tackled the occlusion problem by employing a recurrent neural network trained on human-human interaction data to predict the elbow position during dressing. Rather than relying on vision alone, previous works also explored capacitive proximity sensing \cite{erickson2018tracking} for posture tracking.

In this paper, we will demonstrate that tracking is made easy by hand-holding. With the hand position always known, arm posture tracking can be formulated as an inverse kinematic problem that has analytical solutions and is solvable in real-time with proprioceptive sensors. 

\begin{figure*}[t]
    \centering
    \includegraphics[width=0.8\textwidth]{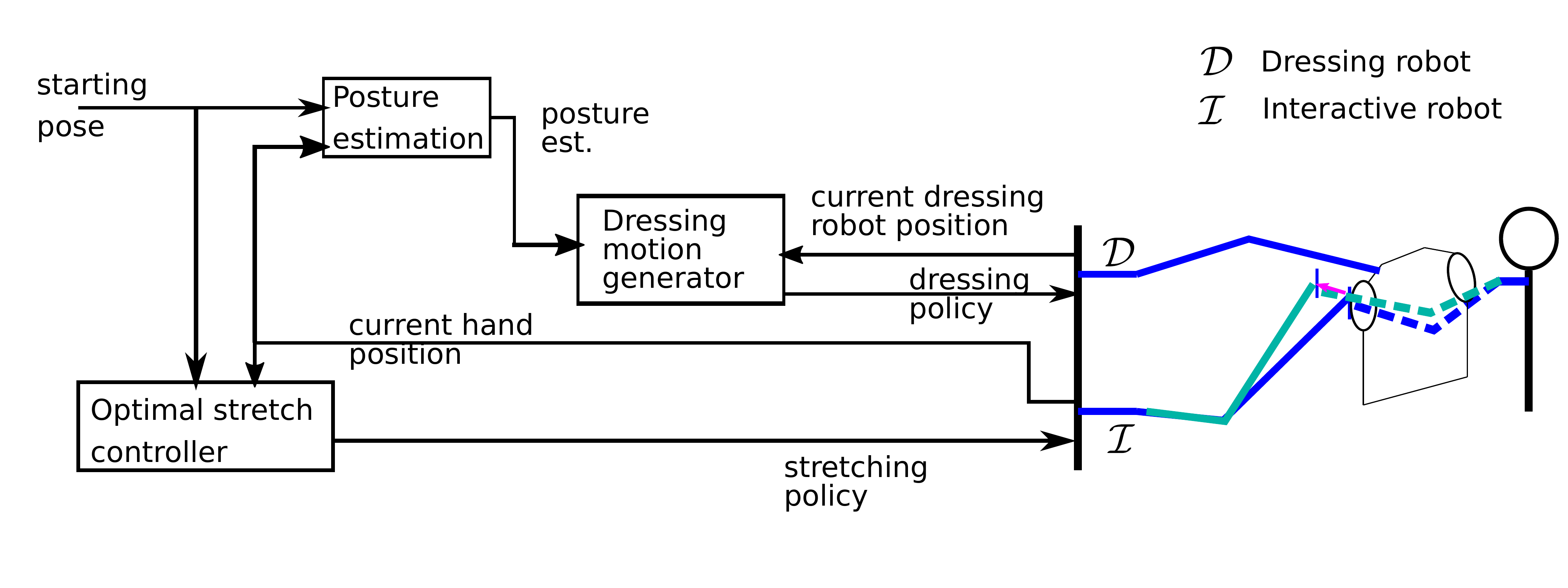}
    \caption{The schematic diagram of cooperative dressing framework. The framework contains three blocks: a posture estimation, a dressing motion generator, and an optimal stretch controller. The posture estimation takes in an initial posture of the human arm and the current hand position and outputs the estimated current arm posture. The posture is then fed into the dressing motion generator together with the current dressing robot position to derive the next step dressing policy. The optimal stretch controller takes the initial human arm posture and the current hand posture yields the stretching policy for the interactive robotics arm.}
    \label{fig:overall_scheme}
\end{figure*}
\section{Framework Overview}\label{sec:framework}
We envision a bimanual dressing assistance system that:
\begin{enumerate}
    \item provides support/guidance to the human arm so as to facilitate dressing,
    \item tracks the arm posture without additional sensors (vision/force/proximity),
    \item can be programmed easily without expert knowledge of robotics,
    \item and flexible to arms of different lengths.
\end{enumerate}

Fig.~\ref{fig:overall_scheme} presents the overall cooperative dressing scheme that fulfills the above goals. The framework contains three parts: an optimal stretch controller, a posture estimation, and a dressing motion generator.

The optimal stretch controller is designed by analysing dressing behaviours under different arm postures to facilitate dressing. The optimal stretch controller takes in the initial human arm posture together with the current hand position and yields the stretching policy for the interactive robot (see Sect.~\ref{sec:optimal_stretch}). 

Benefiting from the hand holding, the real-time posture tracking is done by the posture estimation module which takes in the same inputs as the stretch controller and outputs the estimated current arm posture (see Sect.~ \ref{sec:human_posture_est}). It relies solely on proprioceptive sensors.

Using the estimated posture, we define a posture-dependent dressing coordinate and transform the robot movement in the Cartesian coordinate to the dressing coordinate for learning the dressing motion generator from human demonstrations. One of the benefits of employing LfD is that the motion can be programmed by non-expert users \cite{schmidts2011imitation, scherzinger2019contact, ravichandar2020recent}. The defined dressing coordinate enables flexible encoding of the dressing strategy to adapt to different arm lengths. Inputs to the dressing motion generator are the estimated posture together with the current dressing robot position. Using the inputs, it generates the next step dressing policy (see Sect.~\ref{sec:dressing_robot}).

The scheme described above is realized with a bimanual robot setup illustrated in Fig.~\ref{fig:dressing_setup_overview}. The dressing robot grips the cloth and executes the dressing policy. The interactive robot is equipped with a SoftHand from qbrobotics that provides human-like hand-holding. It supports and guides the human in the dressing process.

In addition, we employ the following assumptions:
\begin{itemize}
    \item The human hand is already in the sleeve opening
    \item The starting arm posture is known,
    \item The shoulder remains static during dressing.
    \item The framework considers dressing one arm only. 
\end{itemize}
\section{Interactive robot -- Optimal Stretch}\label{sec:interactive_robot}
In this section, we motivate a key feature in the human arm posture that affects the dressing policy. Based on this feature, we design an optimal stretch controller for the interactive robot. Finally, we present a scheme for the estimation of human arm posture during dressing.
\subsection{The feature affects dressing -- elbow angle}\label{sec:elbow_angle}
The dressing gets complicated around the elbow. If we simplify the forearm and the upper arm as two straight lines, then the discontinuity between these two lines is the linking point (elbow) which connects both lines. The degree of discontinuity can be represented by the elbow angle $\psi \leq \pi$. The angle $\psi$ is a pure geometrical feature that influences the dressing policy.

If we neglect force and only consider geometry, dressing resembles a hotwire game. The cloth can be simplified into a ring that represents the armhole (more formally referred to as the armscye in textile literature), and the arm is the wire that the ring needs to pass through. The goal of dressing is to transport the ring to a position above the shoulder. 

\begin{figure}[t]
    \centering
    \includegraphics[width=0.48\textwidth]{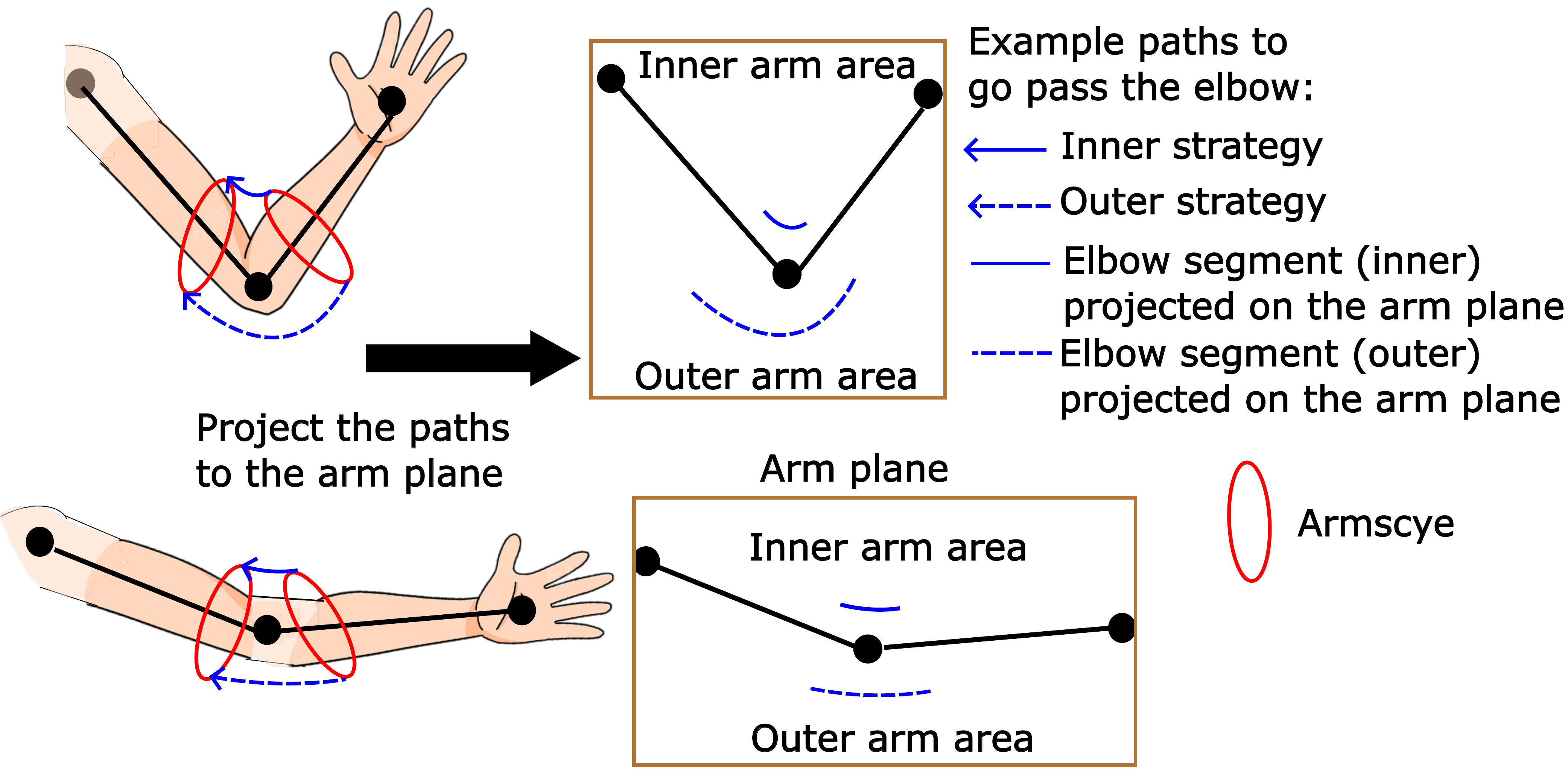}
    \caption{Illustration of the inner and outer path of the armscye. We consider the arm plane defined by the hand, elbow, and shoulder. The arm plane is divided into two parts: the inner arm area with an angle less than $\pi$ and the outer area with an angle more than $\pi$.}
    \label{fig:inner_outer_path}
\end{figure}

However, a major difference that complicates the dressing task is the nature of the ring. In a hotwire game, the ring is rigid. All points on the ring move simultaneously with the grasping/control point on the ring. In dressing, due to the deformation nature of the armscye, we observe the property of diminishing rigidity \cite{berenson2013manipulation} in the armscye --  i.e.\ that the effect of gripper motion along the armscye diminishes as the distance from the gripper increases. This property will affect our dressing strategy when the elbow angle is small. We will explain the reason with Fig.~\ref{fig:inner_outer_path} in the following paragraph.

In Fig.~\ref{fig:inner_outer_path}, we define the arm plane: the plane given by shoulder, elbow, and hand positions. The plane is divided by the arm into two parts, inner arm area (angle between the forearm and upper arm less than $\pi$) and outer arm area (angle between the forearm and upper arm larger than $\pi$). Then we project the dressing path onto the arm plane. 
The path can be divided into three segments. Two are segments that go past the forearm and upper-arm. Another segment is the elbow segment that goes past the elbow and connects the forearm and upper-arm paths. We are interested in the elbow segment of the path which is shown in normal and dashed blue lines in Fig.~\ref{fig:inner_outer_path}. Using the projected paths, we distinguish two different strategies\footnote{The trivial case of the shoulder, elbow, and hand is in a line: we cannot define a plane anymore, yet there is also no difference between an inner and outer arm area.}:
\begin{itemize}
    \item inner strategy: the projection of the elbow segment on the arm plane is within the inner arm area. This strategy offers a shorter path to go around the elbow
    \item outer strategy: the projection of the elbow segment on the arm plane is within the outer arm area. This strategy offers a longer path to go around the elbow.
\end{itemize}

When $\psi$ is small,  taking the inner strategy as shown in blue lines in Fig.~\ref{fig:inner_outer_path}, the far end of armscye where we have very little control needs to travel much longer to go pass the elbow (See Fig.~\ref{fig:inner_outer_path} the dash blue lines). There is a high possibility it fails and results in the armscye getting stuck. It is more plausible in this case to take the outer strategy which travels in the outer arm area, although travels longer in distance, it makes sure that the part with the least control is required to move as little as possible to minimize the chance of getting stuck. When $\psi$ is larger, the distance difference between an inner and outer strategy diminishes and becomes zero when the arm is fully stretched: $\psi = \pi$.  

To conclude, elbow angle $\psi$ affects the dressing strategy. When $\psi$ is large, the degree of discontinuity becomes small (the arm resembles a straight line), and we are more likely to take the inner strategy to have a smaller path length. When $\psi$ is small, the degree of discontinuity gets large (the arm is more bend), and we will more likely to take the outer strategy to avoid getting stuck. We will show this effect with expert demonstration data on different static arm postures (with different elbow angles) in Sect.~\ref{sec:exp_effect_elbow}.
\subsection{Optimal stretch based on the elbow angle}\label{sec:optimal_stretch}
The outer strategy requires a longer dressing trajectory, and when $\psi$ is small, there is always the risk of getting stuck even taking the outer strategy. Thus, the objective of the interactive robot is to guide the human arm so that $\psi$ increases as much as possible to avoid getting stuck and facilitate dressing.

\begin{figure}[t]
    \centering
    \includegraphics[width=0.35\textwidth]{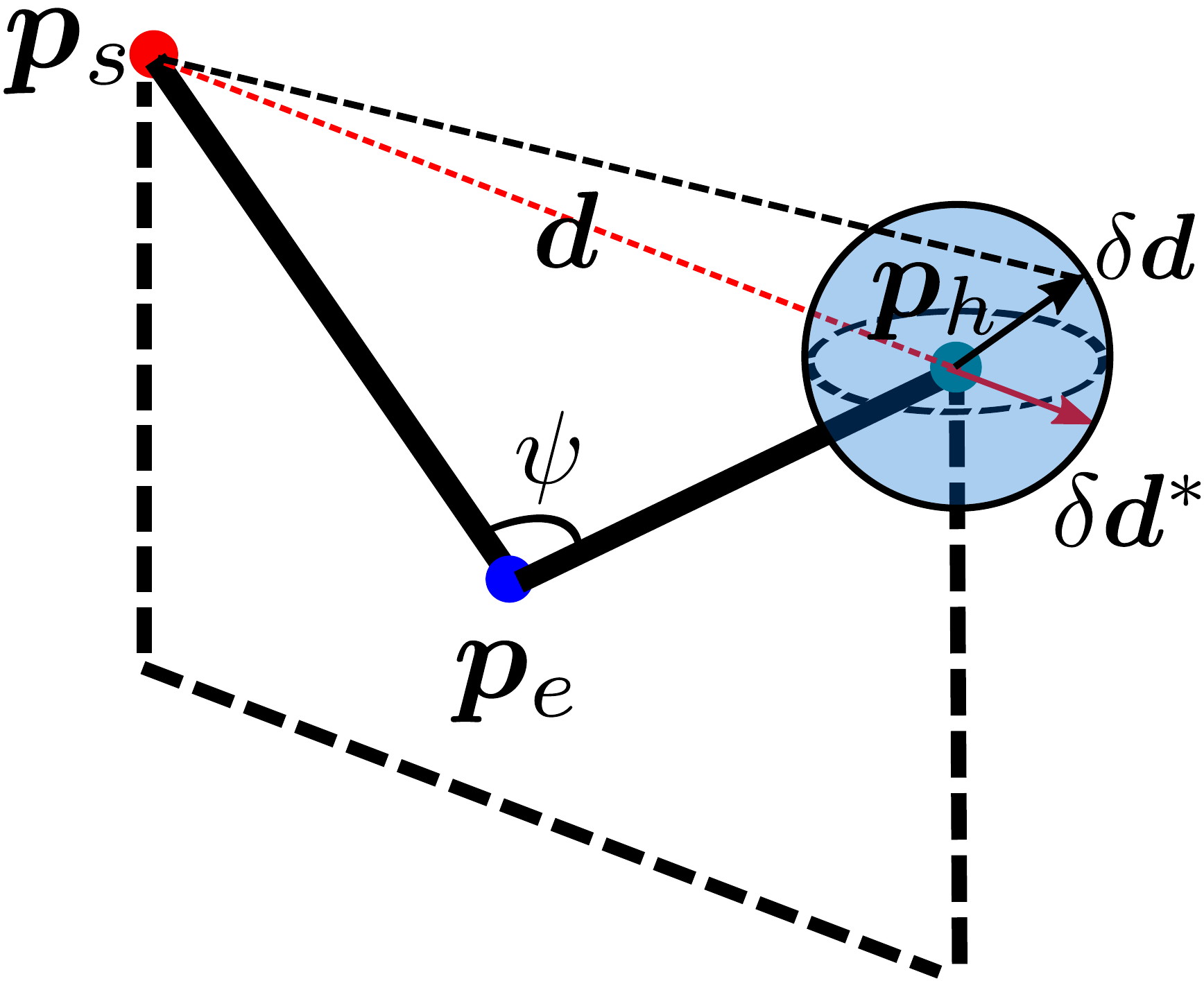}
    \caption{The optimal direction that maximizes the increase of the elbow angle $\psi$. The arm posture is given by the shoulder, elbow, and hand joint positions: $\vP = \{\vp_{s}, \vp_{e}, \vp_{h}\}$. The optimal direction $\delta\vd^*$ is aligned with the direction that connects the hand and the shoulder and marked with a red solid arrow. The triangle inequality theorem is demonstrated with another random direction $\delta\vd$ and the resulting shoulder-to-hand distance is marked with a black dash line.}
    \label{fig:optimal_stretch_direction}
\end{figure}

The interactive robot is holding the human hand. Besides the singular configuration where the arm is fully stretched, the robot is able to guide the hand to move in all directions in 3D which is represented by the blue sphere in Fig.~\ref{fig:optimal_stretch_direction}. In the figure, the arm posture is given by the shoulder, elbow, and hand joint positions: $\vP = \{\vp_{s}, \vp_{e}, \vp_{h}\}$. The movement direction $\delta \vd$ that maximizes the increase of the elbow angle, here referred to as $\delta\vd^*$, is aligned with the direction of $\overrightarrow{\vp_{s}\vp_{h}}$. Proving the optimal direction is straightforward, since $\psi \propto |\vd + \delta\vd|$, finding $\delta \vd^* $ that maximizes $\psi$ is equivalent to:
\begin{equation*}
    \delta\vd^* = \max_{\delta\vd} |\vp_{s}\vp_{h} + \delta \vd|,~\text{for}~ \delta \rightarrow 0, ~\text{and}~ \vd \in \mathrm{O}(3)
\end{equation*}

With the triangle inequality theorem, we have:
\begin{equation*}
    |\vp_{s}\vp_{h} + \delta \vd| \leq |\vp_{s}\vp_{h}| + |\delta \vd| 
\end{equation*}
The maximum value for $ |\vp_{s}\vp_{h} + \delta \vd|$ is when the equality holds. And it only holds when $\delta \vd$ and $\vp_{s}\vp_{h}$ are in the same direction.

Therefore, to increase the elbow angle as much as possible, the guidance force direction from the interactive robot should always be aligned with the direction that connects the hand and the shoulder.

Knowing the optimal direction to move, we further implement the optimal stretch controller with Cartesian impedance control. 
The dynamic equation of the robot can be modeled according to
\begin{equation}
    \bm{{M}}(\bm{q}) \ddot{\bm{q}}+\bm{ {C}}(\bm{q},\dot{\bm{q}})+\bm{ {G}}(\bm{q})= \bm{\tau}_{task} + \bm{\tau}_{ext} 
\end{equation}
where, in order from left to right, there are the mass, the Coriolis and the gravitational term that depend on the joint configuration $\bm{q}$ and, on the right, the torque for the Cartesian (or task) control and the externally applied torque. 

The input to the low-level controller is the desired cartesian position (and orientation) obtained from the dressing policy. A PD control rule is used to generate the desired control force in the end effector simulating the impedance behavior. This is then converted into a control joint torque using the equivalent principle $ \bm{\tau}  = \bm{J}^{\top} \bm{F}_{cart}$.  

Using the PD control rule and the force/torque equivalence principle, the task space torque is defined as
\begin{equation*}
    \bm{\tau}_{task}=\bm{{J}}^\top \left[ \bm{ {K}}(\bm{x}_{d}-\bm{x})+\bm{D} (\dot{{\bm{x}}}_{d}-\dot{{\bm{x}}}) \right]+\bm{ {C}}(\bm{q},\dot{\bm{q}})+\bm{ {G}}(\bm{q})
\end{equation*}
where $\bm{J}$ is the geometric Jacobian, and the stiffness $ \bm{K}$ and the damping $ \bm{D}$ give the compliant behaviour with a critically damped response; $\bm{x}_{d}$ and $\dot{{\bm{x}}}_{d}$ are the desired robot cartesian position and velocity; $\bm{x}$ and $\dot{{\bm{x}}}$ are the current cartesian position and velocity of the manipulator. Please notice that the gravitational and the Coriolis terms are compensated to avoid having undesired behaviors on the simulated impedance, such as having a constant offset in the gravity direction.  
In our implementation, we set the desired velocity to zero. 

Given a rotation matrix $\vGamma$ of a local stiffness matrix with respect to the global coordinate, the equivalent stiffness matrix in global coordinates is 
\begin{equation}
    \vK_\text{global}= \vGamma ^T \vK_\text{local} \vGamma
\end{equation}
where in our framework, the rotation matrix $\vGamma$ transforms the $x$-axis of the global coordinates in the direction of end effector movements that maximizes the increase of the human elbow angle $\psi$, i.e.\ $\overrightarrow{\vp_{s}\vp_{h}}$.

Additionally, we choose
\[
\vK_\text{local} = \begin{bmatrix}
k_x & 0 & 0 \\
0 & 0 & 0 \\
0 & 0 & 0
\end{bmatrix},
\] 
in order to have a non-zero stiffness only in the direction of movement and complete compliance in the other two perpendicular directions: such a controller will provide a guiding force in the optimal direction of hand movement that maximizes the increase of the elbow angle $\psi$. At the same time, the human can disagree and move easily in the plane perpendicular to the force direction.




\subsection{Human arm posture estimation}\label{sec:human_posture_est}
With the guidance force upon the hand, the human arm posture is changing during dressing. For successfully dressing the human arm, we propose a real-time vision-free estimation of the human arm posture.

Consider two coordinate frames $\mathcal{O}^\mathcal{D}$, $\mathcal{O}^\mathcal{I}$ located at the base of each robot respectively. Any point $\vx^\mathcal{I}$ in $\mathcal{O}^\mathcal{I}$ can be transformed into $\mathcal{O}^\mathcal{D}$ by:
\begin{equation}\label{eq:transformation}
    \vx^{D} = \vR\vx^{I} + \vt.
\end{equation}
where $\vR$ is a $3$ by $3$ rotation matrix and $\vt$ is the translation vector. We can obtain $\vR$ and $\vt$ between two coordinates by calibration. 

The human hand is joined by the interactive robot $\mathcal{I}$. Thus its position can be regarded as the same as the end-effector position of the interactive robot $\vx^{I}_{\text{end-eff}}$. Using (\ref{eq:transformation}), we can track the human hand position in $\mathcal{O}^\mathcal{D}$ given $\vR$ and $\vt$:
\begin{equation}
    \vp_h^\mathcal{D} = \vR\vx^{I}_{\text{end-eff}} + \vt. 
\end{equation}

Assuming the initial arm posture in $\mathcal{O}^\mathcal{D}$ is known and the shoulder does not move during the dressing, we can move the origin of $\mathcal{O}^\mathcal{D}$ to the shoulder by translation, we denote this frame as shoulder frame. From now on, for simplicity, unless explicitly denoted with superscript $^\mathcal{I}, ^\mathcal{D}$, all positions are with regard to the shoulder frame with its origin at the shoulder.

\begin{figure}[b]
    \centering
    \includegraphics[width=0.4\textwidth]{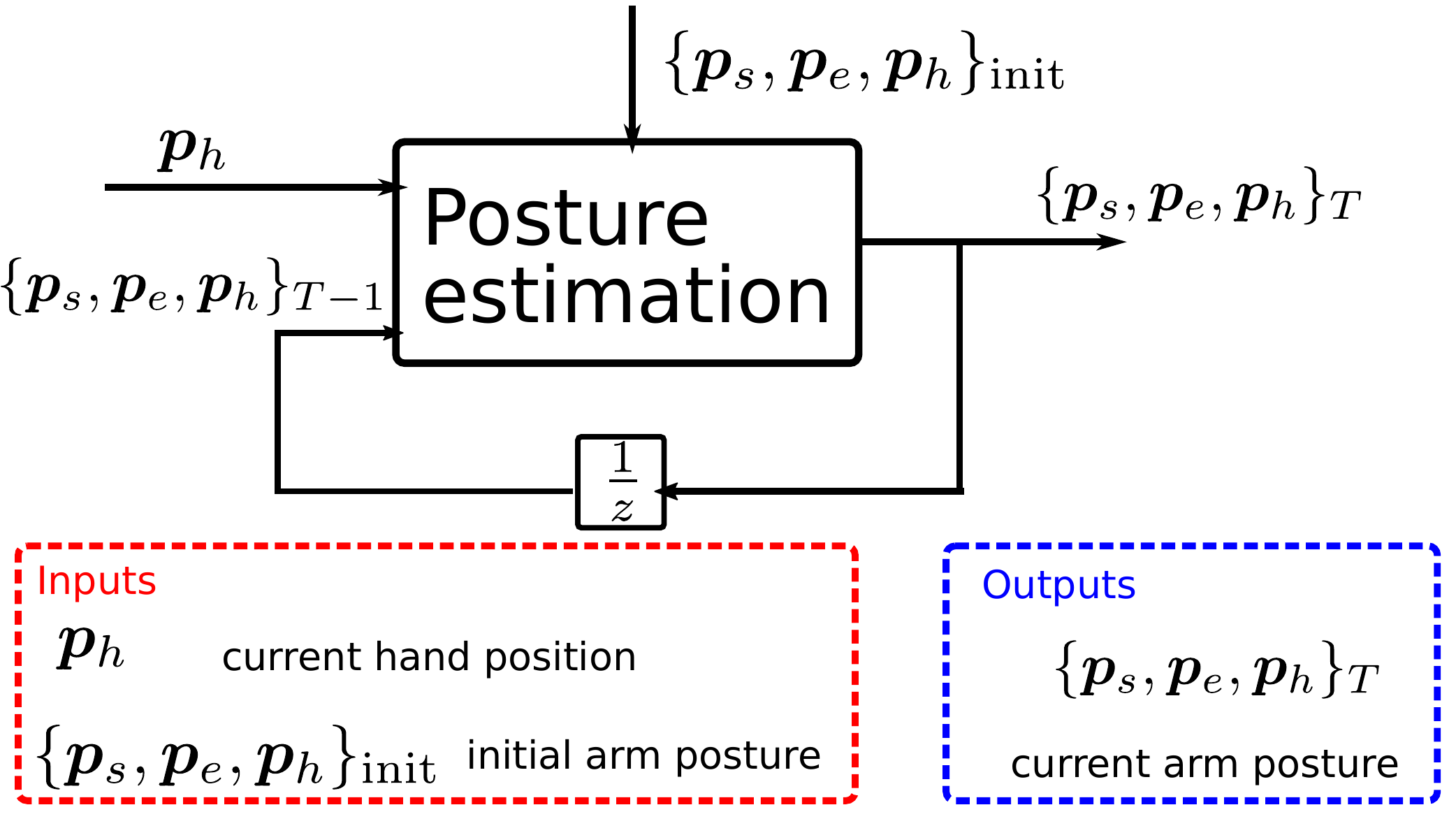}
    \caption{A schematic diagram showing the inputs and outputs of the human posture estimation.}
    \label{fig:arm_posture_est}
\end{figure}

Since the shoulder is assumed static during dressing, the posture estimation problem boils down to recovering the elbow position $\vp_{e}$ given the hand position $\vp_h$. A schematic diagram is shown in Fig.~\ref{fig:arm_posture_est}.

\begin{figure}[t]
    \centering
    \includegraphics[width=0.4\textwidth]{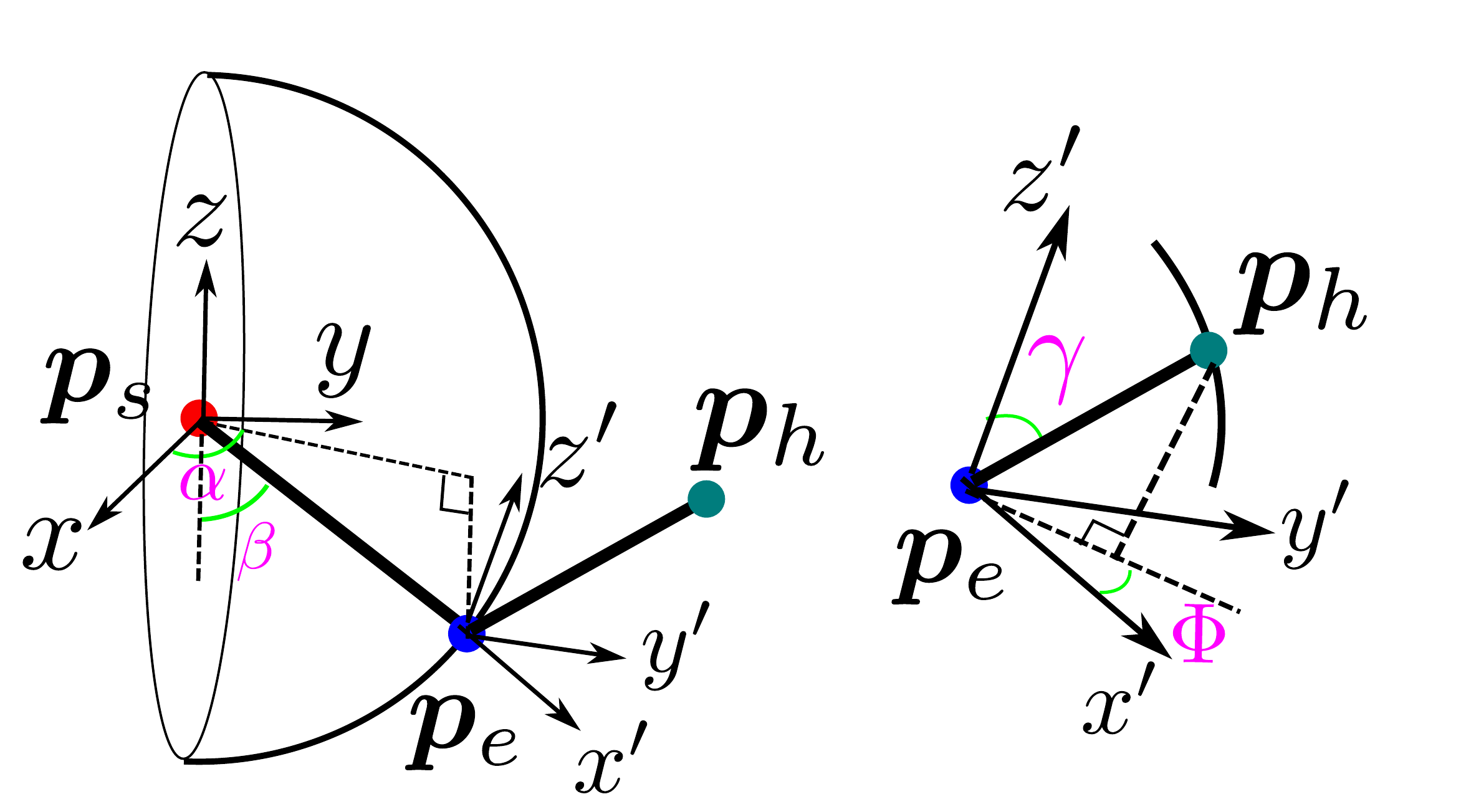}
    \caption{Schematic representation of human arm from the shoulder to the hand joint with four joint angles. A spherical joint located at the shoulder with two degrees of freedom ($\alpha$, $\beta$),  and another spherical joint at the elbow with two degrees of freedom ($\phi$, $\gamma$).}
    \label{fig:human-arm-joint-representation}
\end{figure}

The human arm has seven degrees of freedom. Only four out of seven contribute to the displacement of the hand $\vp_h$ \cite{desmurget1997postural}. The four joint angles $\vq = [\alpha~\beta~\phi~\gamma]^T$ are depicted in Fig.~\ref{fig:human-arm-joint-representation}. A spherical joint located at the shoulder with two degrees of freedom ($\alpha$, $\beta$), and another spherical joint at the elbow with two degrees of freedom ($\phi$, $\gamma$). The shoulder, elbow, and hand representation in the shoulder frame can be transformed into the angle representation and vice versa (given forearm and upper arm lengths). The hand position can be expressed as:
\begin{equation}
    \vp_h = \vf(\vq)
\end{equation}
The hand movement:
\begin{equation}
    \vp_h + \Delta \vp_h = \vf(\vq + \Delta \vq) = \vf(\vq) + \vJ(\vq) \Delta \vq + \text{h.o.t.},
\end{equation}
where 
\begin{equation}
    \vJ(\vq) = \frac{\partial \vf}{\partial \vq}(\vq) \in \mathbb{R}^{3 \times 4}
\end{equation}
This then leads to:
\begin{equation}\label{eq:delta_IK}
    \Delta \vp_h = \vJ(\vq) \Delta \vq.
\end{equation}
For a given $\Delta \vp_h$, (\ref{eq:delta_IK}) has infinite solutions of $\Delta \vq$. We aim to find the $\Delta \vq$ from (\ref{eq:delta_IK}) with good accuracy as compared to the human movement while enabling fast computation to be implemented in real-time. Thus, we define a quadratic optimization with an analytical solution:  
\begin{equation}\label{eq:optim_ik_arm_posture}
    \begin{aligned}
        & \min_{\Delta \vq} \vL = \Delta \vq^T \vQ \Delta \vq \\
        & \text{s.t. }\Delta \vp_h = \vJ(\vq) \Delta \vq,
    \end{aligned}
\end{equation}
where $\vQ$ is a diagonal weighting matrix with non-zero diagonal elements for solving the optimal $\Delta \vq$. Physically, $\vQ$ indicates how the hand velocity is distributed on each joint, a larger element on the diagonal indicates the corresponding joint is less likely to move during arm movements. We find $\vQ$ using the arm movement data recorded during the stretch and the detail is presented in Sect.~\ref{sec:eval_posture_est}.

The solution to the optimization problem (\ref{eq:optim_ik_arm_posture}) is (detailed derivation in the Appendix):
\begin{equation*}
     \Delta \vq^* = \vJ^+ \Delta \vp_h - ({\vmu^T \vQ \Delta\vq_h}/{\vmu^T \vQ \vmu}) \vmu,
\end{equation*}
where $\vJ^+$ is Moore–Penrose inverse of the $\vJ$ and $\vmu$ is the null space vector of $\vJ$\footnote{Detail solution can be found in the appendix}. Current postures in angle representation are then:
\begin{equation}
    \hat{\vq}_T = \hat{\vq}_{T-1} + \Delta \vq^*.
\end{equation}
Note that this is a recursive estimation as depicted in Fig.~\ref{fig:arm_posture_est}. Only the initial value of $\vq$ is known such that $\hat{\vq}_{0} = \vq_0$. The analytical solution can be computed fast online which allows us to use the estimation scheme in real-time during dressing. 

Lastly, the angle representation $\hat{\vq}_T$ is transformed into current positions of the shoulder, elbow, and hand $\{\hat{\vp}_s, \hat{\vp}_e, \hat{\vp}_h\}_T$ for deriving the dressing policy. We evaluate the accuracy of the estimation scheme in Sect.~\ref{sec:eval_posture_est}.

\section{Dressing Robot -- Dressing Coordinate-based  Learning}\label{sec:dressing_robot}
The dressing is highly dependent on the human arm posture. Previous research explored such dependency in learning the dressing policy. In \cite{clegg2018learning}, the authors placed in simulation haptic sensor on each segment of the arm as observation in deep reinforcement learning. Others define task parameters as local coordinates at the shoulder, elbow, and hand and then encode the dressing policy with TP-GMM \cite{hoyos2016incremental, pignat2017learning, zhu2022learning}. Similarly, we explore such dependencies and define an arm-posture-dependent coordinate system for encoding the dressing policy with LfD. In the following subsections, we first present the dressing coordinate then the LfD in the new coordinate system. 
\subsection{Dressing coordinate}\label{sec:dressing_coord}
The dressing motion follows the arm from the hand to the elbow and finally reaches the shoulder. The monotonic motion naturally leads us to introduce a scalar quantity to model the dressing progress. We denote the progress scalar as $s$.

\begin{figure}[b]
    \centering
    \subfloat[]{\includegraphics[width=0.6\columnwidth]{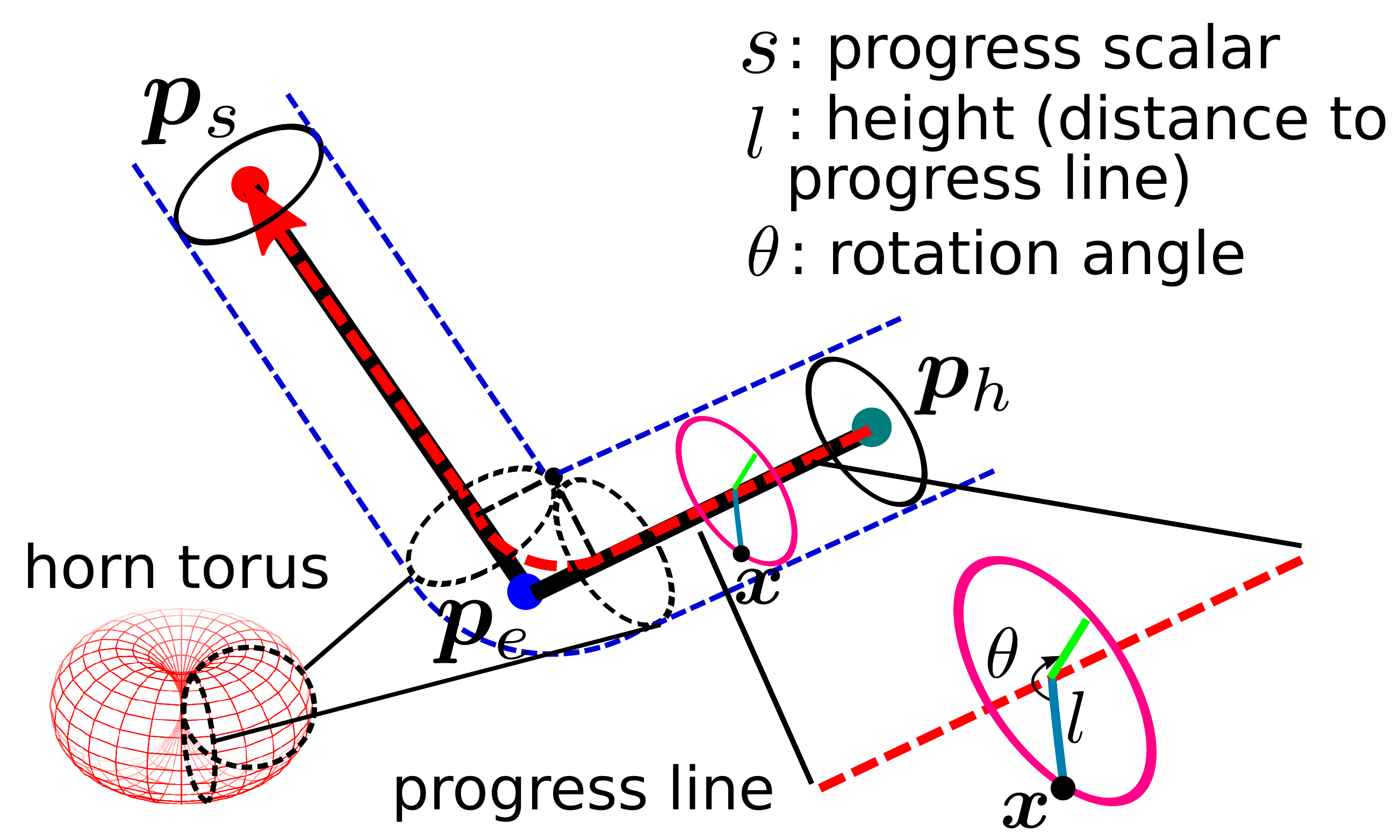}\label{fig:dressing_coord_def}}
    \subfloat[]{\includegraphics[width=0.4\columnwidth]{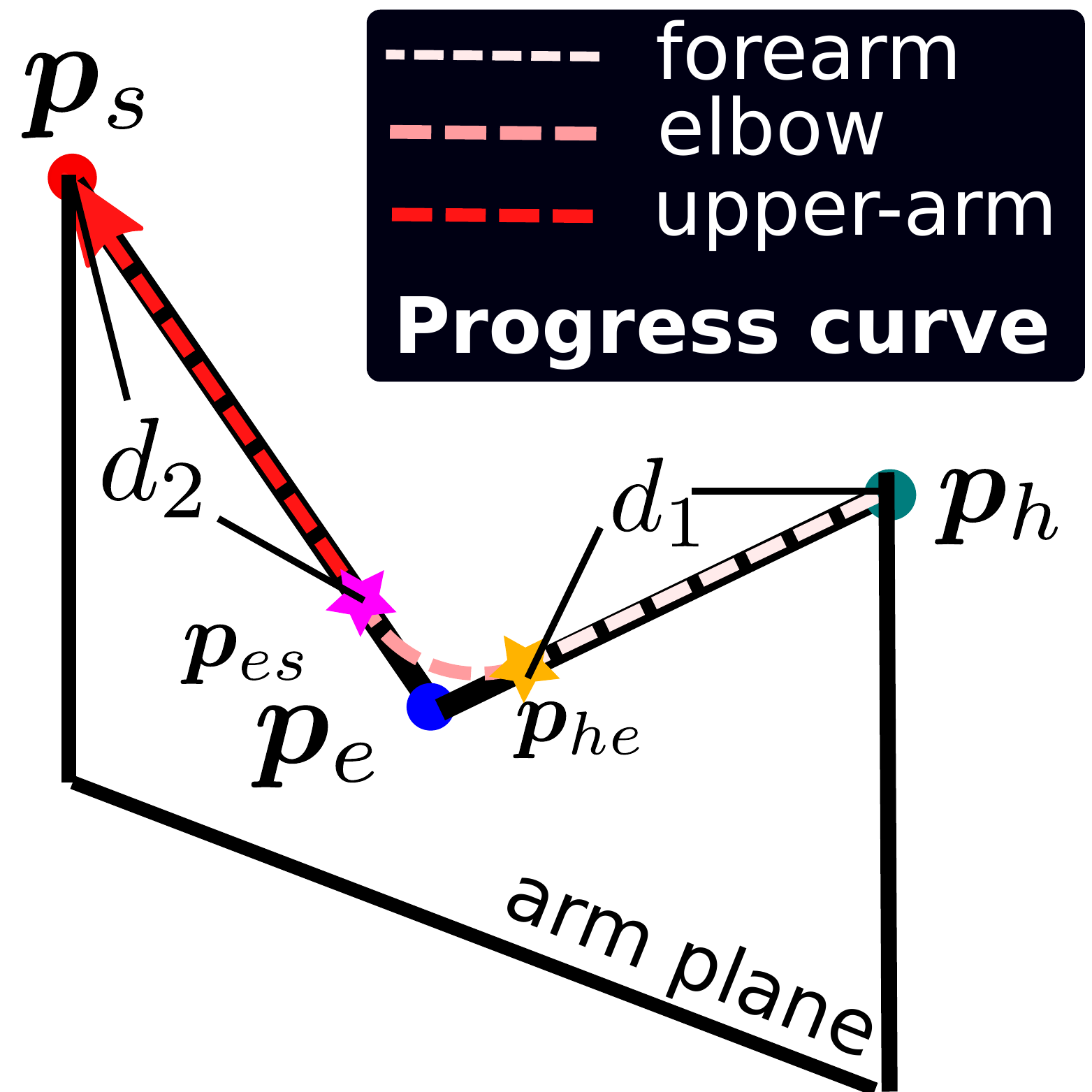}\label{fig:progress_curve}}
    \caption{The dressing coordinate is a cylinder coordinate defined around the arm, with a smoothing transitional corner around the elbow. The sub-figure (a) presents the full definition of the coordinate and (b) shows the progress curve in three segments.}
\end{figure}

The dressing motion is additionally constrained by the size of the armscye. Thus we use a polar coordinate around the forearm and the upper-arm for encoding the constraint: $l$, the distance to the arm, and $\theta$, the angle around the arm. The formulation yields two cylinder coordinates on the forearm and the upper-arm. However, the complexity arises from the elbow where we require a smooth transition from the forearm coordinate to the upper arm coordinate (see Fig.~\ref{fig:dressing_coord_def}).

To enable a smooth transition, we use a part of a horn torus (A torus where the distance from the center of the tube to the center of the torus equals the radius of the tube) for connecting the two cylinder coordinate. As a result of the torus define around the elbow, the progress curve is divided into three pieces: a forearm segment, an elbow arc for a smooth transition to the upper arm, and an upper-arm segment as shown in Fig.~\ref{fig:progress_curve}. 

To convert a point on the dressing path $\vx \in \mathrm{R}^3$ in Cartesian coordinate into the dressing coordinate $[s, l, \theta]$, we first project $\vx$ onto the arm plane given by the shoulder $\vp_s$, elbow $\vp_e$ and hand positions $\vp_h$:
\begin{equation}
    \vx_{\text{arm}} = \vx + \frac{\vv (\vp_{h} \cdot \vv - \vx \cdot \vv)}{|\vv|^2},
\end{equation}
where $\vv$ is the vector perpendicular to the arm plane. 
 
We further find the progress scalar of $\vx_{\text{arm}}$ on the progress curve. The progress curve is divided into $3$ segments by $2$ transitional points: $\vp_{he}$ (transition from the forearm to elbow) and $\vp_{es}$ (transition from elbow to upper arm) as shown by stars on Fig.~\ref{fig:progress_curve}. The distance from $\vp_h$ to $\vp_{he}$ is denoted $|d_1|$, and from $\vp_s$ to $\vp_{es}$ is $|d_2|$.

We need to determine which segments $\vx_{\text{arm}}$ belongs to. To do that, we compute the distance $\vx_{\text{arm}}$ project onto the line defined by the forearm and upper-arm segments respectively:
\begin{align}
    \begin{split}
        d_{\text{forearm}} & = \frac{(\vx_{\text{arm}} - \vp_h) \cdot (\vp_e - \vp_h)}{|(\vp_e - \vp_h)|} \\
        d_{\text{upperarm}} & = \frac{(\vx_{\text{arm}} - \vp_s) \cdot (\vp_e - \vp_s)}{|(\vp_e - \vp_s)|} 
    \end{split}
\end{align}

The value of $d_{\text{forearm}}$ and $d_{\text{upperarm}}$ is then compared against $d_1$ and $d_2$ so that:
\begin{equation*}
  \vx_{\text{arm}} \text{ on the }
    \begin{cases}
      \text{forearm} & \text{if } d_{\text{forearm}} < |d_1| \text{ and } d_{\text{upperarm}} > |d_2|\\
      \text{upperarm} & \text{if } d_{\text{upperarm}} < |d_2| \text{ and } d_{\text{forearm}} > |d_1|\\
      \text{elbow} & \text{otherwise}
    \end{cases}       
\end{equation*}

\begin{figure}[!thbp]
    \centering
    \includegraphics[width=0.4\textwidth]{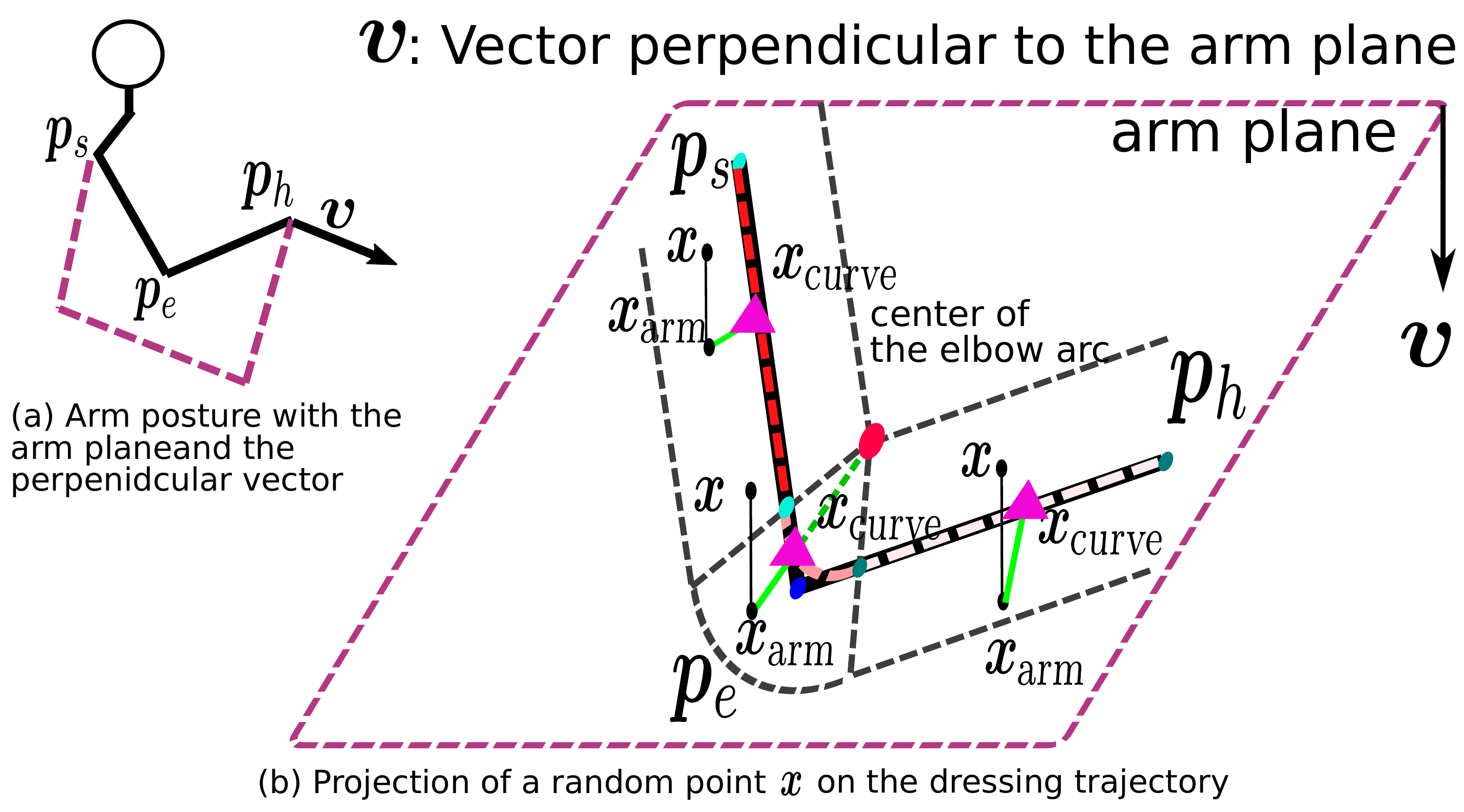}
    \caption{Graphic illustration of finding the corresponding point on the progress curve $\vx_{\text{curve}}$ given a random point $\vx$ on the dressing path in Cartesian coordinate. The point $\vx$ is first projected onto the arm plane to get $\vx_{\text{arm}}$. Then we determine which segment $\vx_{\text{arm}}$ belongs to on the progress curve. Finally we find the corresponding $\vx_{\text{curve}}$.}
    \label{fig:convert_to_dressing_coord}
\end{figure}

Depending on which segment $\vx_{\text{arm}}$ belongs to, we find the corresponding $\vx_{\text{curve}}$ (pink triangles on Fig.~\ref{fig:convert_to_dressing_coord}) on the progress curve for $\vx_{\text{arm}}$. For the forearm and upper arm segment, calculating $\vx_{\text{curve}}$ is straightforward. For the elbow segment, we find the line connecting the center of the elbow arc to $\vx_{\text{arm}}$. The intersection point between the line and the progress curve is $\vx_{\text{curve}}$ (See Fig.~\ref{fig:convert_to_dressing_coord}).

We then define $s = 0$ for $\vx_{\text{curve}} = \vp_h$, $s = 1$ for $\vx_{\text{curve}} = \vp_s$, for any $\vx_{\text{curve}}$, we have:
\begin{equation*}
  s = 
    \begin{cases}
        \vspace{0.8mm}
      \frac{d_{\text{forearm}}}{|d_{\text{curve}}|} & \text{if $\vx_{\text{curve}}$ is on the forearm} \\
      
     \frac{|d_{\text{curve}}| - d_{\text{upperarm}}}{|d_{\text{curve}}|} & \text{if $\vx_{\text{curve}}$ is on the upperarm} \\
      \vspace{0.8mm}
      \frac{d_1 + \omega r}{|d_{\text{curve}}|} & \text{if $\vx_{\text{curve}}$ is on the upperarm}
    \end{cases}       
\end{equation*}
where $|d_{\text{curve}}|$ is the full length of the progress curve from $\vp_h$ to $\vp_s$, $\omega$ is the angle of $\vx_{\text{curve}}$ on the elbow arc and $r$ is the radius of the elbow segment on the progress curve.

Once we found the projection of the dressing position on the progress curve $\vx_{\text{curve}}$ and the progress scalar $s$, we can calculate the distance to the arm $l$ and the angle around the arm $\theta$ as follows.

The length $l$ of $\vx$ is:
\begin{equation}
    l = |\vx - \vx_{\text{curve}}|
\end{equation}

We define $\vv$ as the reference $0$\textdegree~for the angle $\theta$ computation. The $\theta$ is defined as
\begin{equation}
    \theta = \angle(\vx_{\text{curve}} - \vx,\vv)
\end{equation}
Note that $\vv$ has two directions, one pointing towards to the human body and the other pointing outwards. We select the former to avoid flipping angles from $0$ to $2\pi$ during the dressing motion.  

By converting the Cartesian coordinate to the dressing coordinate and obtaining the motion generation model in the dressing coordinate, we can:
\begin{itemize}
    \item limit the policy space to be around the arm by imposing the maximum value for the length $l$, 
    \item ensure the motion convergence to the vicinity of the shoulder by progress scalar $s$,
    \item transfer the trained motion generation model on one arm to another as the data is relative to the arm posture.
\end{itemize}
Note one prerequisite for the last statement is that the ratio between the forearm and upper arm is similar among people.

\subsection{LfD in dressing coordinate}\label{sec:lfd_dressing}
For LfD in dressing coordinate, we employ Gaussian Mixture Models (GMM) and Gaussian Mixture Regression (GMR) for learning and generating the dressing policy \cite{calinon2007learning}. The GMM/GMR-based imitation learning offers fast movement retrieval and versatile inputs/outputs arrangement which is desirable for our task. Please note that the LfD algorithm is not a part of our contribution and it is possible to use different methods/formulations for learning the policy in the dressing coordinate. 

We first brief describe GMM/GMR-based LfD then show how to apply it to learning the dressing policy.

GMM models the joint distribution of the demonstration data. We denote the demonstrated data $\vxi$:
\begin{equation}\label{eq:demo_data_structure}
  \vxi = \begin{bmatrix}
  \vxi^\mathcal{I}, \vxi^\mathcal{O}
  \end{bmatrix}^T.
\end{equation}
with $\vxi^\mathcal{I}$ is the input and $\vxi^\mathcal{O}$ the output. The probability density function $p(\vxi_i)$ is estimated with $K$ Gaussian distributions:
\begin{equation*}
    p(\vxi_i) = \sum_{k=1}^{K}\pi_k \mathcal{N}(\vxi_i \big| \vmu_k, \Sigma_k)
\end{equation*}
where $\vmu_k$ and $\Sigma_k$ are mean and variance of the $k$th Gaussian. To determine an optimal number of Gaussian $K$ for GMM, we make use of the Bayesian Information Criterion (BIC) \cite{schwarz1978estimating} for balancing the model complexity and representation quality. 

Once $K$ is selected, we can initialize the GMM with K-Means clustering and then employ the Expectation-Maximization algorithm to iterative compute the model parameters $\big\{ \pi_k, \vmu_k, \vSigma_k \big\}^K_{k = 1}$.

The resulting $K$ Gaussian parameters can be decomposed as:
\begin{equation}
      \vmu_k = \begin{bmatrix}
      \vmu_k^\mathcal{I} \\
      \vmu_k^\mathcal{O}
      \end{bmatrix}, \quad
      \vSigma_k = \begin{bmatrix}
      \vSigma_k^\mathcal{I} & \hat{\vSigma}_k^{\mathcal{I}\mathcal{O}} \\
      \vSigma_k^{\mathcal{O}\mathcal{I}} & \hat{\vSigma}_k^\mathcal{O}
      \end{bmatrix}.
  \end{equation}
Given input data the output is retrieved using GMR by conditioning on the input data \cite{muhlig2009task}. 

For encoding dressing with GMM/GMR in the dressing coordinate, we demonstrate the dressing task with static arm postures, the arm posture, and the dressing path is recorded for converting the dressing path into the dressing coordinate. 
We use the progress scalar $s$ and the elbow angle $\phi$ as inputs.
The outputs are the differences in the distance to the arm $\delta l$, and the angle difference around the arm $\delta \theta$. The differences are computed with regard to the starting position in the demonstration. For all demonstrations, we start from above the hand of the human. 

\section{Experiments}\label{sec:experiments}
We conduct three different experiments to investigate various aspects of the proposed framework.

In Sect. \ref{sec:exp_effect_elbow}, we investigate the effect of the elbow angle on the dressing strategy using expert demonstration data of static arm postures with different elbow angles.
Then in Sect. \ref{sec:eval_posture_est} we evaluate the arm posture estimation scheme to show that it is accurate enough for the dressing task. 

Lastly, in Sect. \ref{sec:interactive_dressing} we showcase our interactive dressing framework with real robot experiments. To demonstrate the flexibility of the defined dressing coordinate, we train the policy from demonstrations on a mannequin and apply the policy to human arms with different lengths. 
We compared our trajectory encoding method (using dressing coordinates) with TP-GMM based methods \cite{hoyos2016incremental, pignat2017learning, zhu2022learning} in terms of dressing success rate.

Furthermore, we conduct an ablation study where we disable the stretch movement and dress with a static human arm. Additionally, we test the framework under cases where the human is not fully compliant with the interactive robot and show that the overall framework still works well.

In these experiments, the robot dresses the human in short and long sleeves shirts with different materials. Note that the automatic dressing of long sleeves shirts are unprecedented in previous research.

\subsection{Effect of the elbow angle on dressing strategies}\label{sec:exp_effect_elbow}
In Sect.~\ref{sec:elbow_angle} we discussed the effect of the elbow angle on the dressing strategy with diminished rigidity. In this section, we aim to show this effect with expert demonstration data.

\begin{figure}[t]
    \centering
    \subfloat[]{\includegraphics[width=0.15\textwidth]{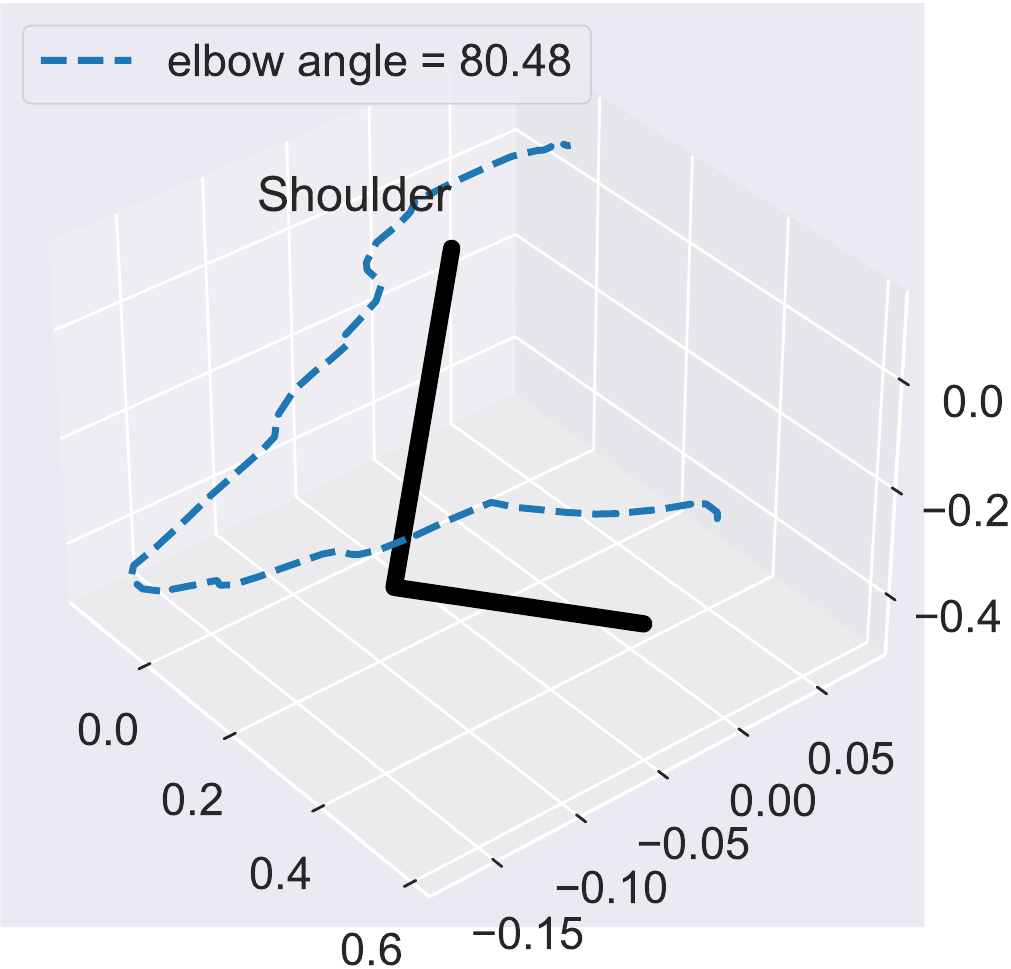}\label{fig:elbow_1}}
    \hspace{0.5mm}
    \subfloat[]{\includegraphics[width=0.145\textwidth]{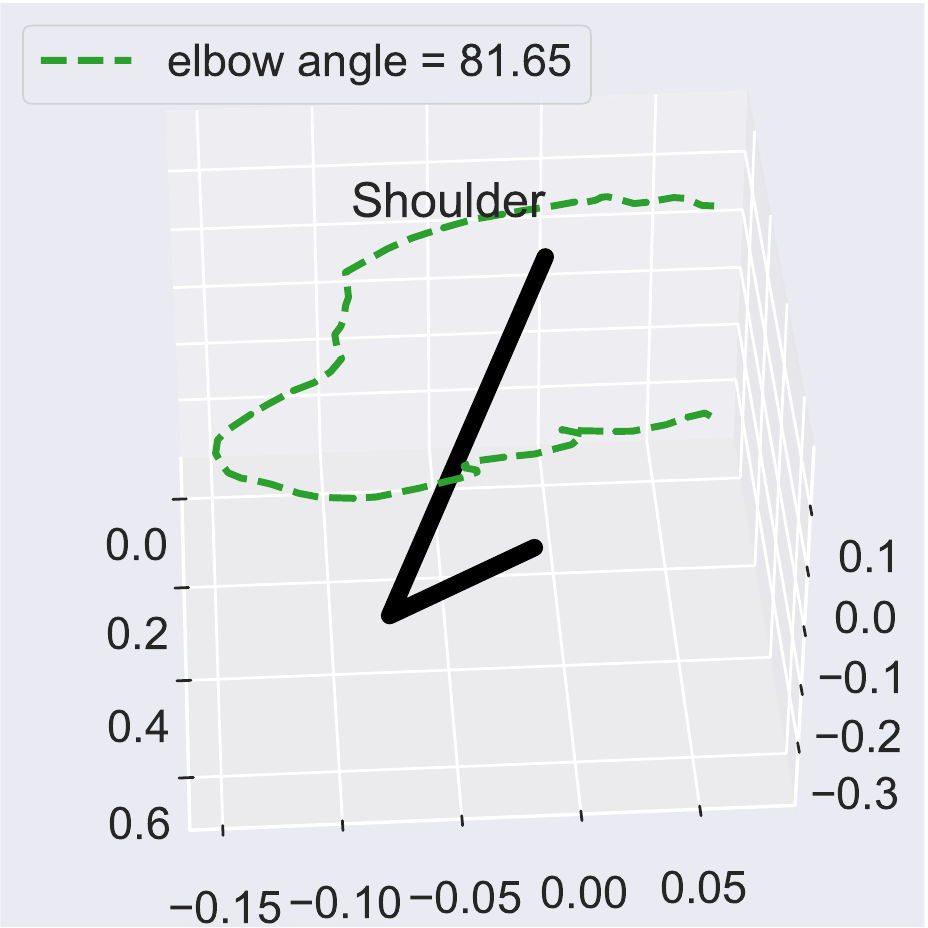}\label{fig:elbow_2}}
    \hspace{0.5mm}
    \subfloat[]{\includegraphics[width=0.15\textwidth]{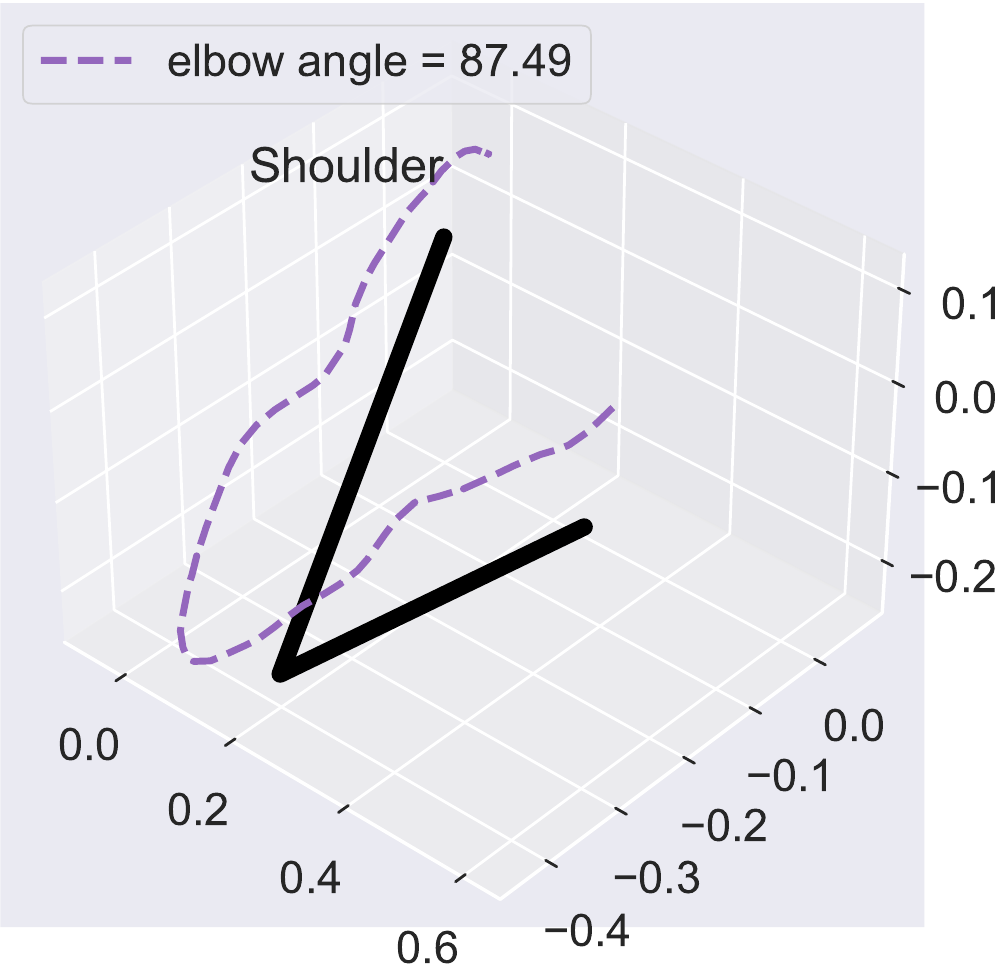}\label{fig:elbow_3}} \\
    \subfloat[]{\includegraphics[width=0.15\textwidth]{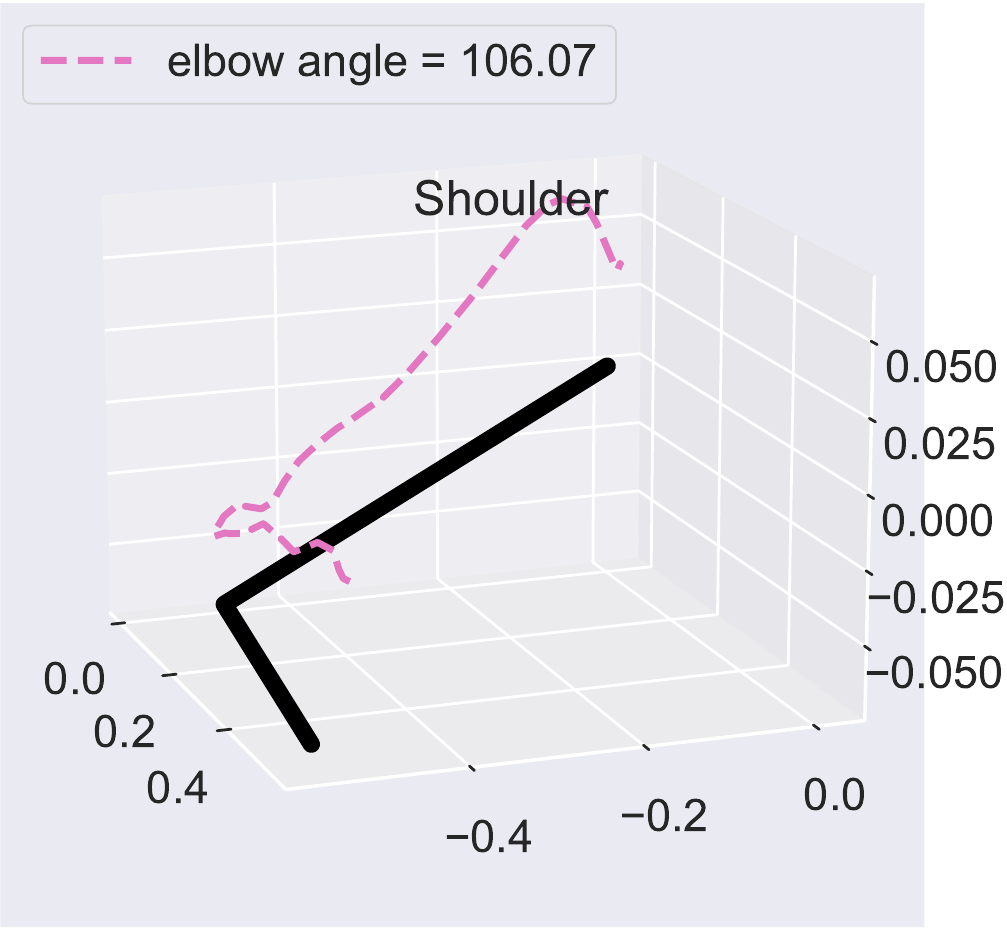}\label{fig:elbow_4}}
    \hspace{0.5mm}
    \subfloat[]{\includegraphics[width=0.15\textwidth]{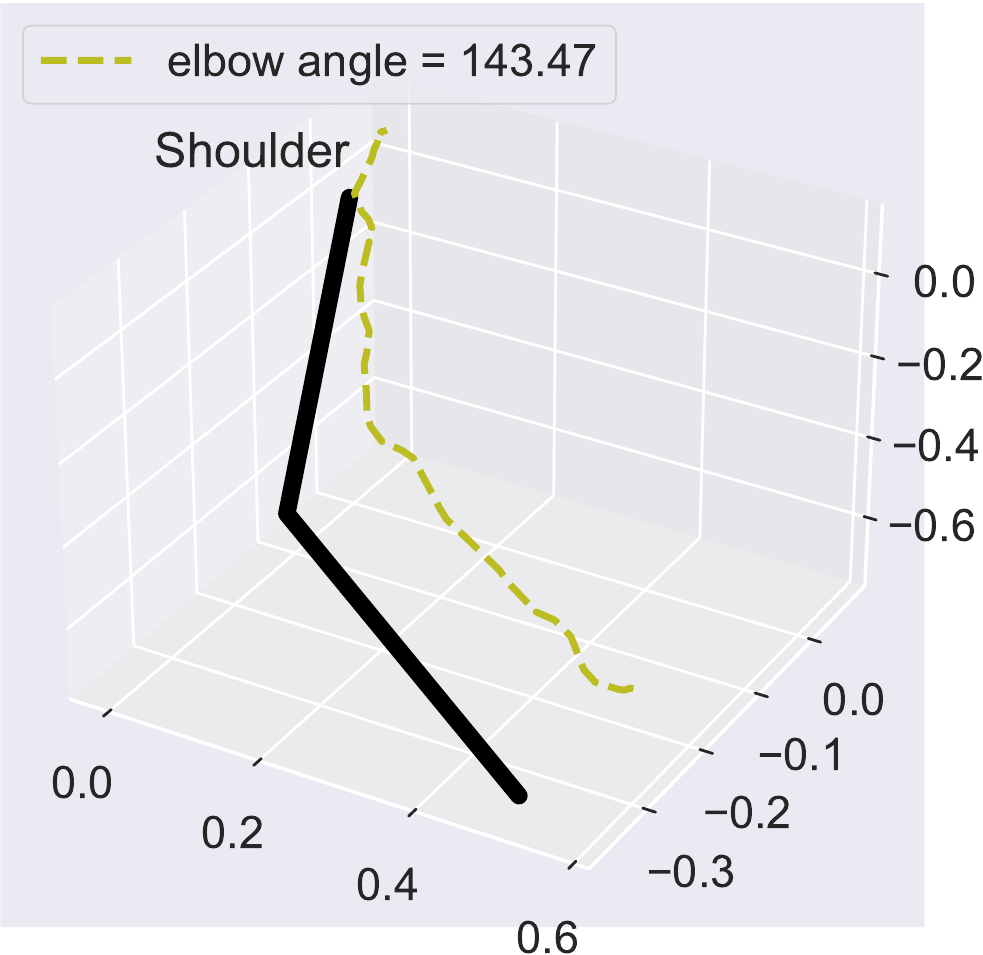}\label{fig:elbow_5}}
    \hspace{0.5mm}
    \subfloat[]{\includegraphics[width=0.15\textwidth]{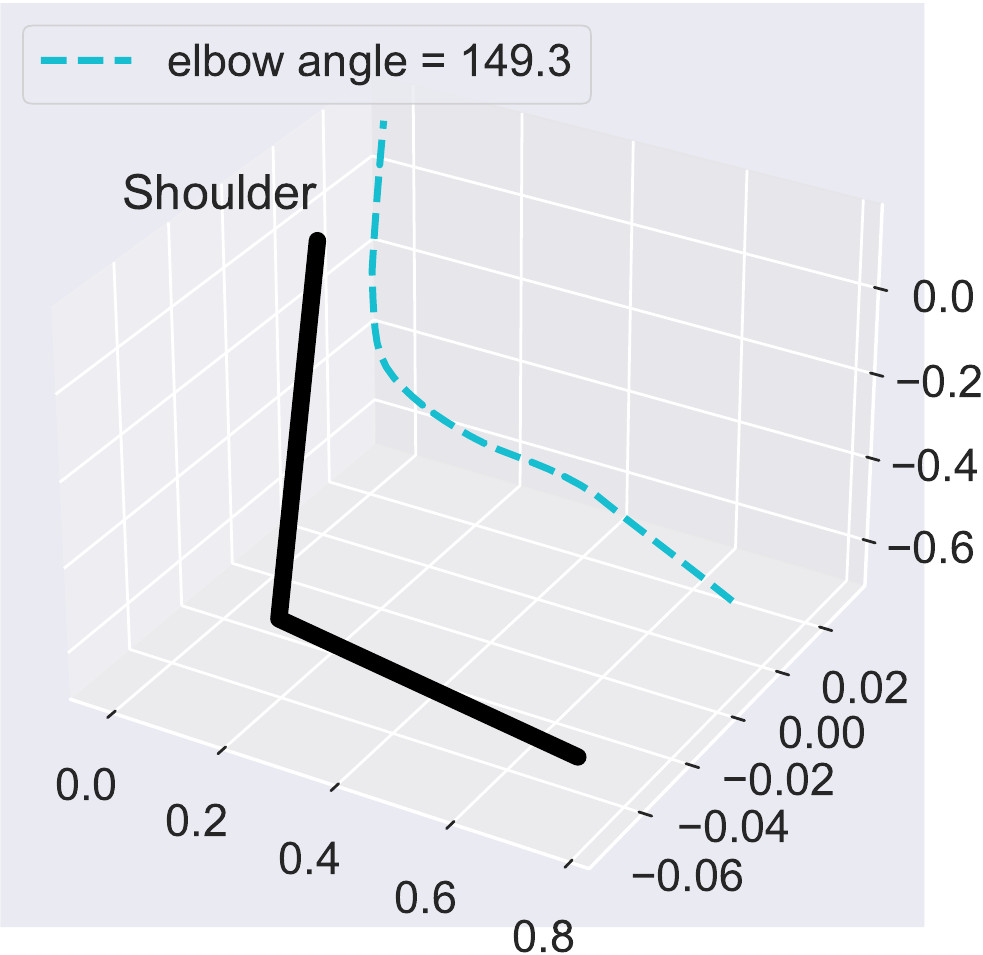}\label{fig:elbow_6}}
    \caption{Static arm postures with different elbow angles and the corresponding expert demonstrations of dressing with these postures.}
    \label{fig:elbow_angle_n_dressing_strategies}
\end{figure}

We collect demonstration data through kinesthetic teaching with a Franka robot under different arm postures. The dressing paths and arm postures are recorded in the process. In Fig.~\ref{fig:elbow_angle_n_dressing_strategies} we present $6$ postures with different elbow angles. We draw each posture and dressing path individually. In each figure, the solid black line is the arm posture where the shoulder positions are at the top and labeled with \say{Shoulder}. The dashed line in each figure is the dressing path that corresponds to the posture.

\begin{figure}[t]
    \centering
    \subfloat[]{\includegraphics[width=0.16\textwidth]{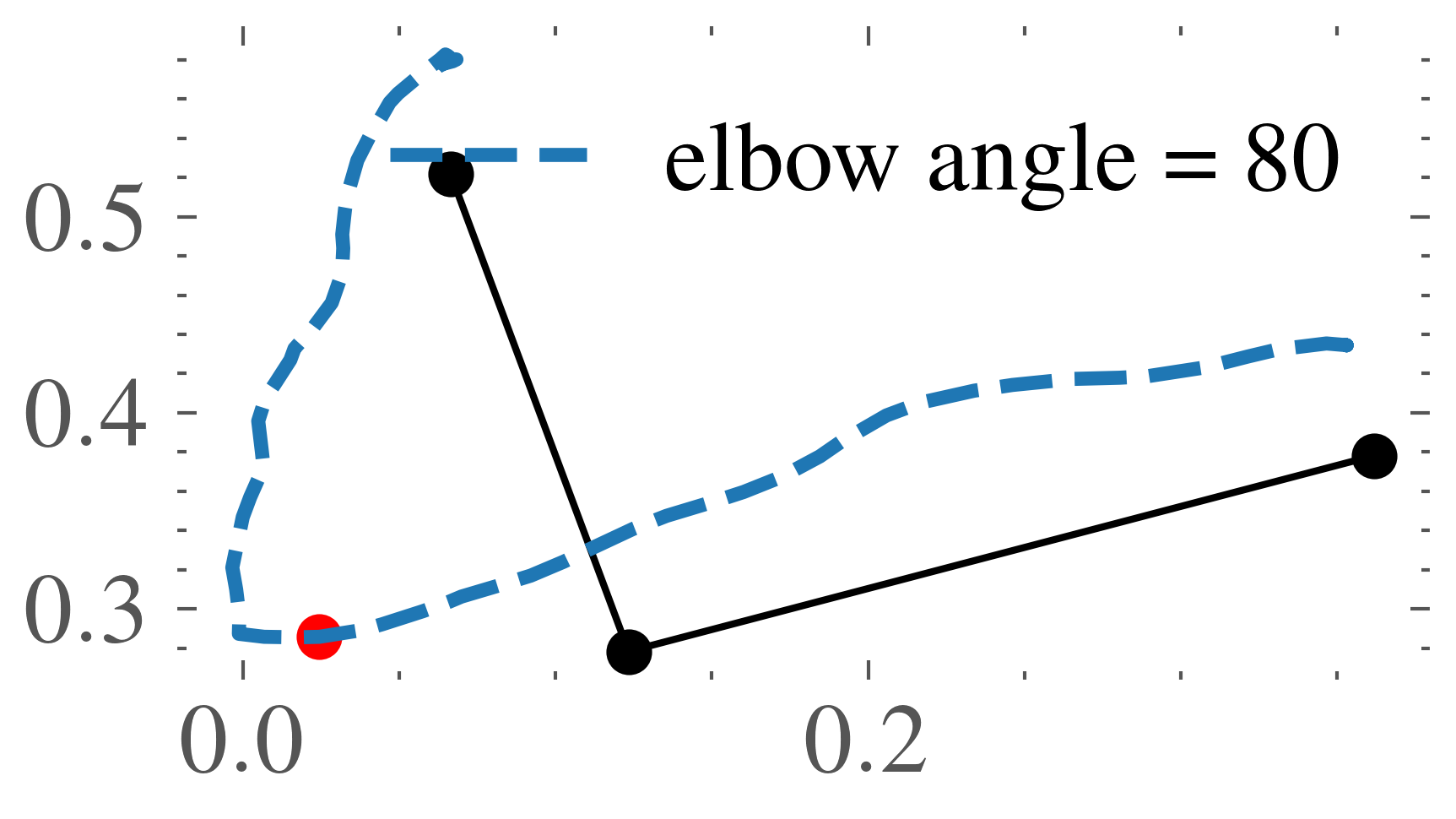}\label{fig:elbow_1}}
    \subfloat[]{\includegraphics[width=0.16\textwidth]{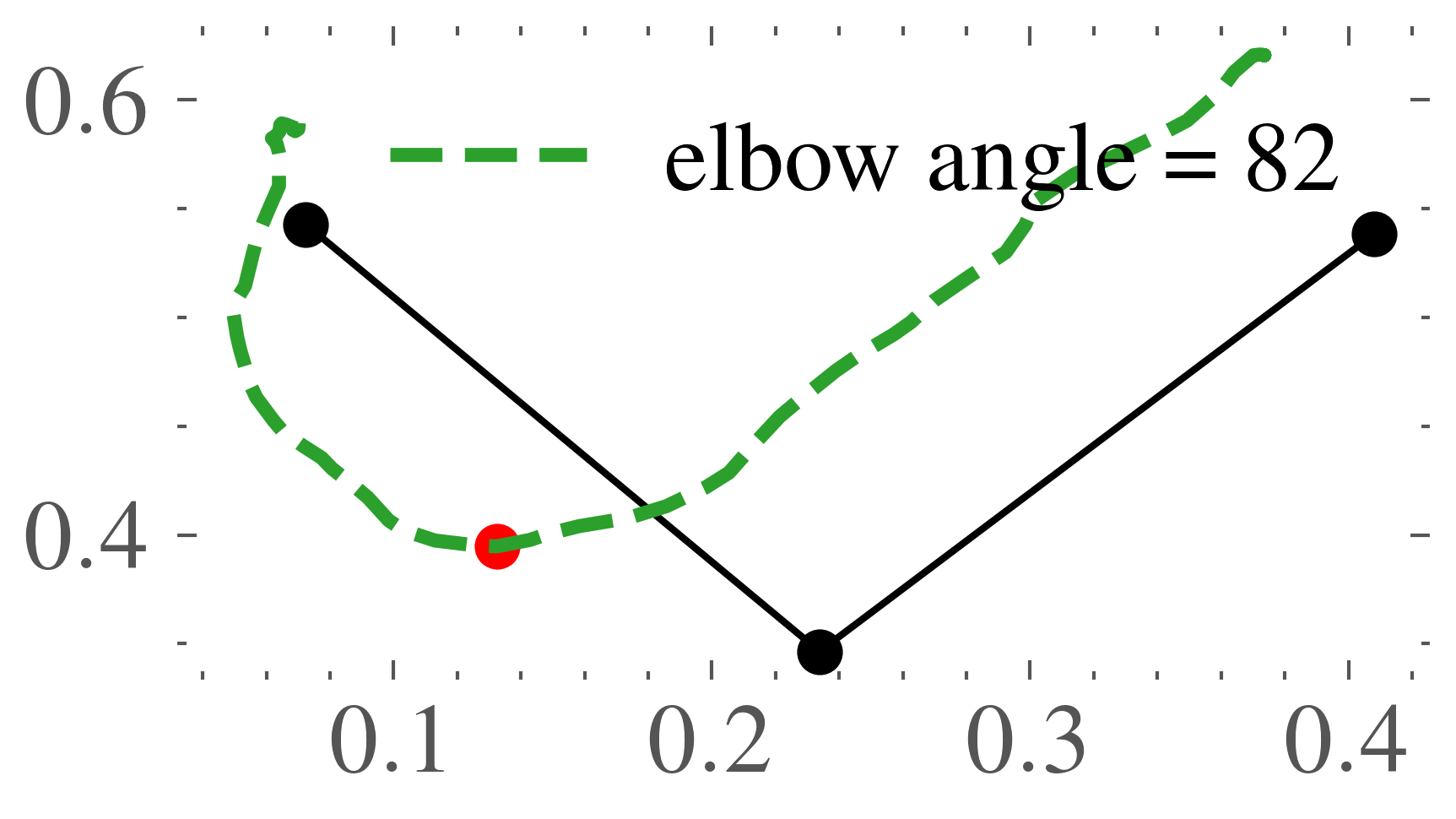}\label{fig:elbow_2}}
    \subfloat[]{\includegraphics[width=0.16\textwidth]{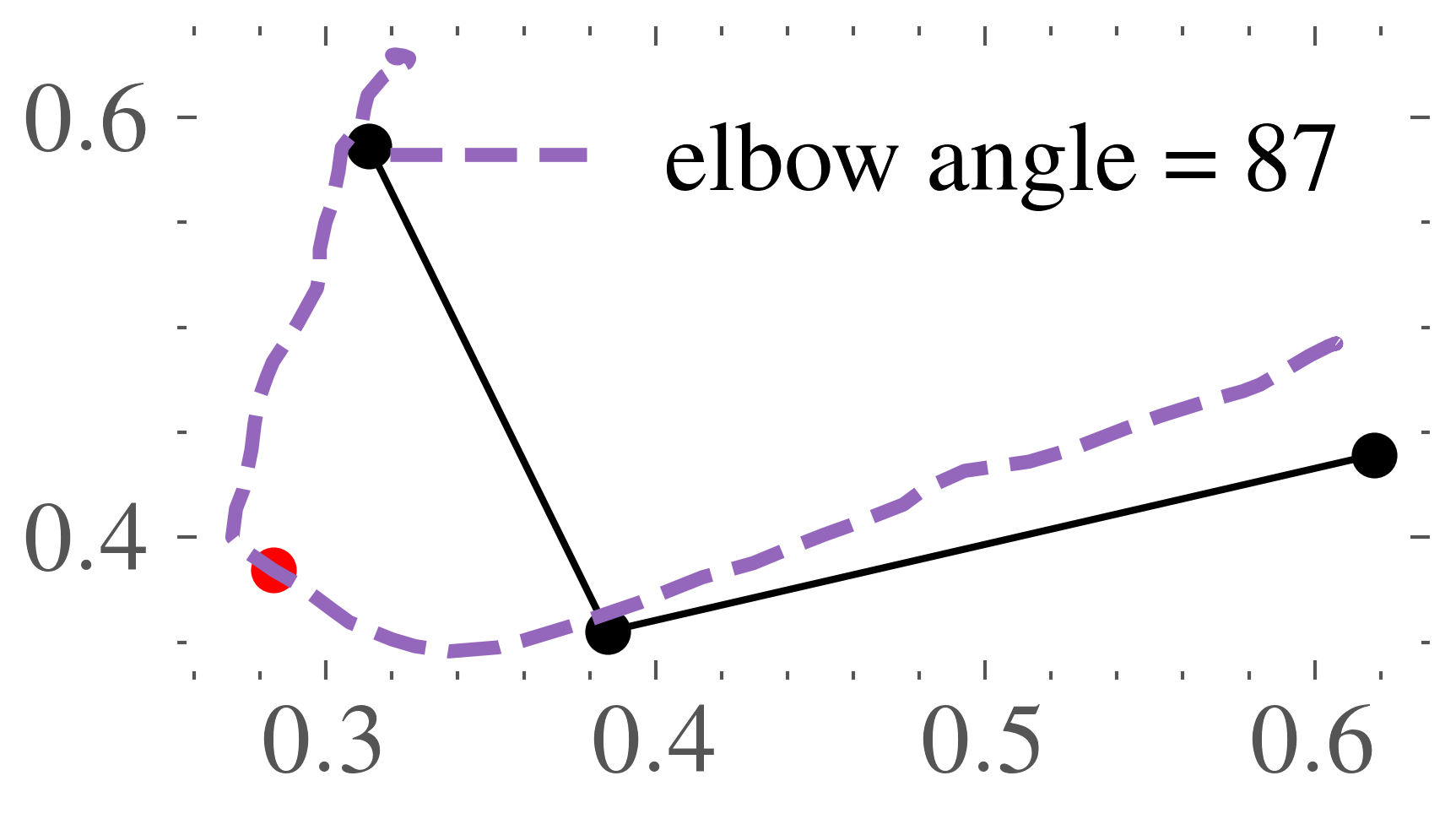}\label{fig:elbow_3}} \\
    \subfloat[]{\includegraphics[width=0.16\textwidth]{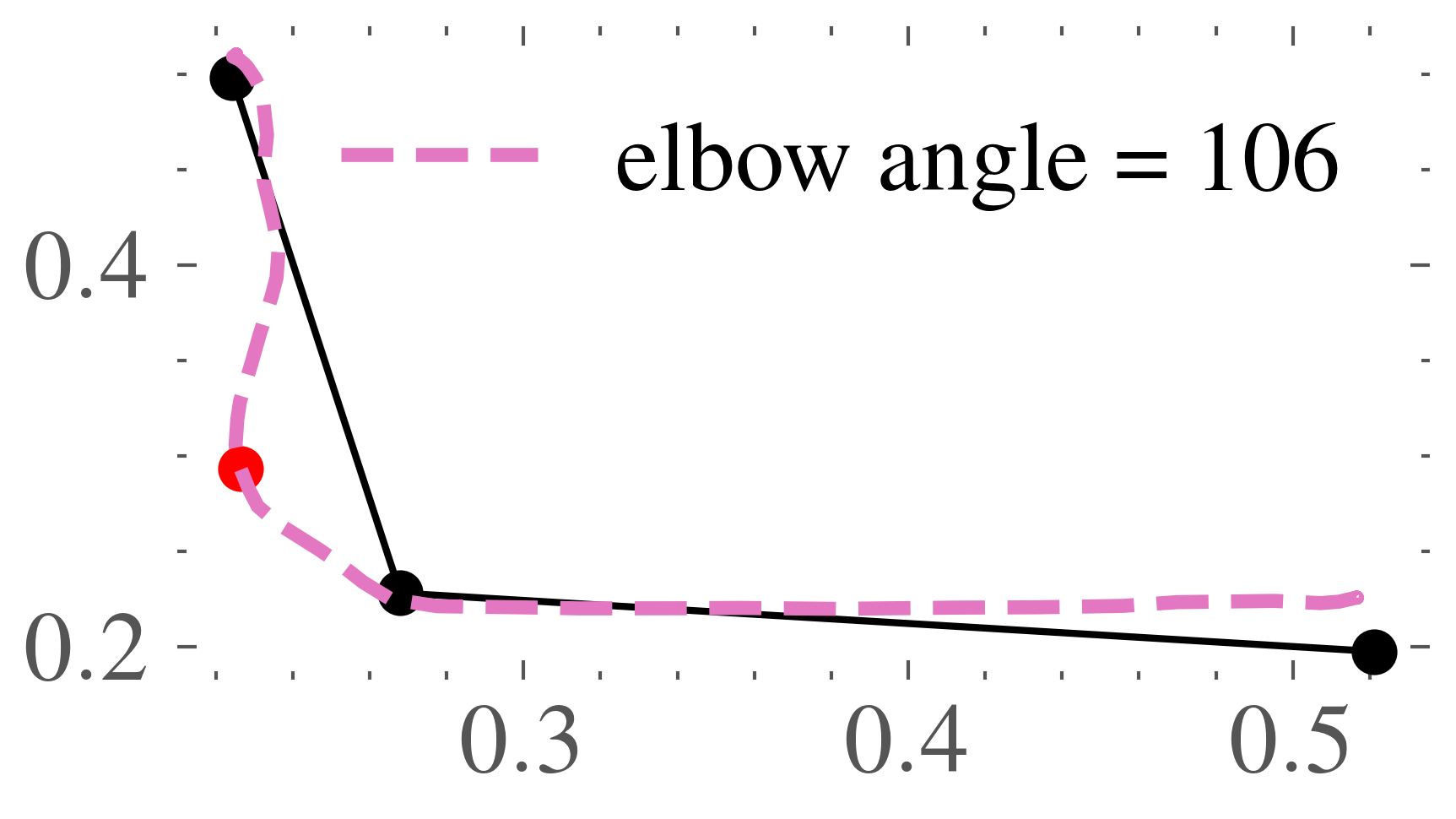}\label{fig:elbow_4}}
    \subfloat[]{\includegraphics[width=0.16\textwidth]{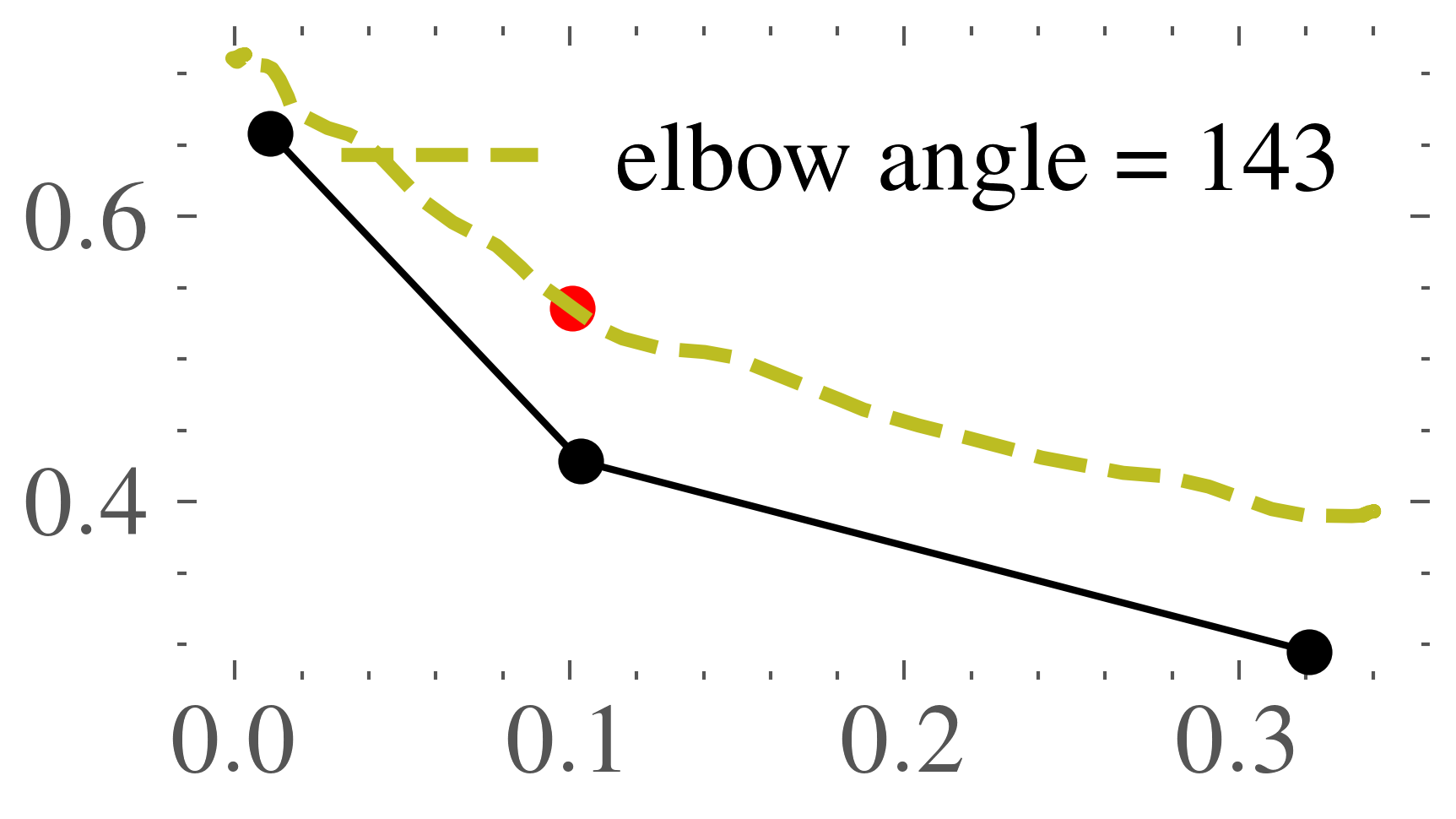}\label{fig:elbow_5}}
    \subfloat[]{\includegraphics[width=0.16\textwidth]{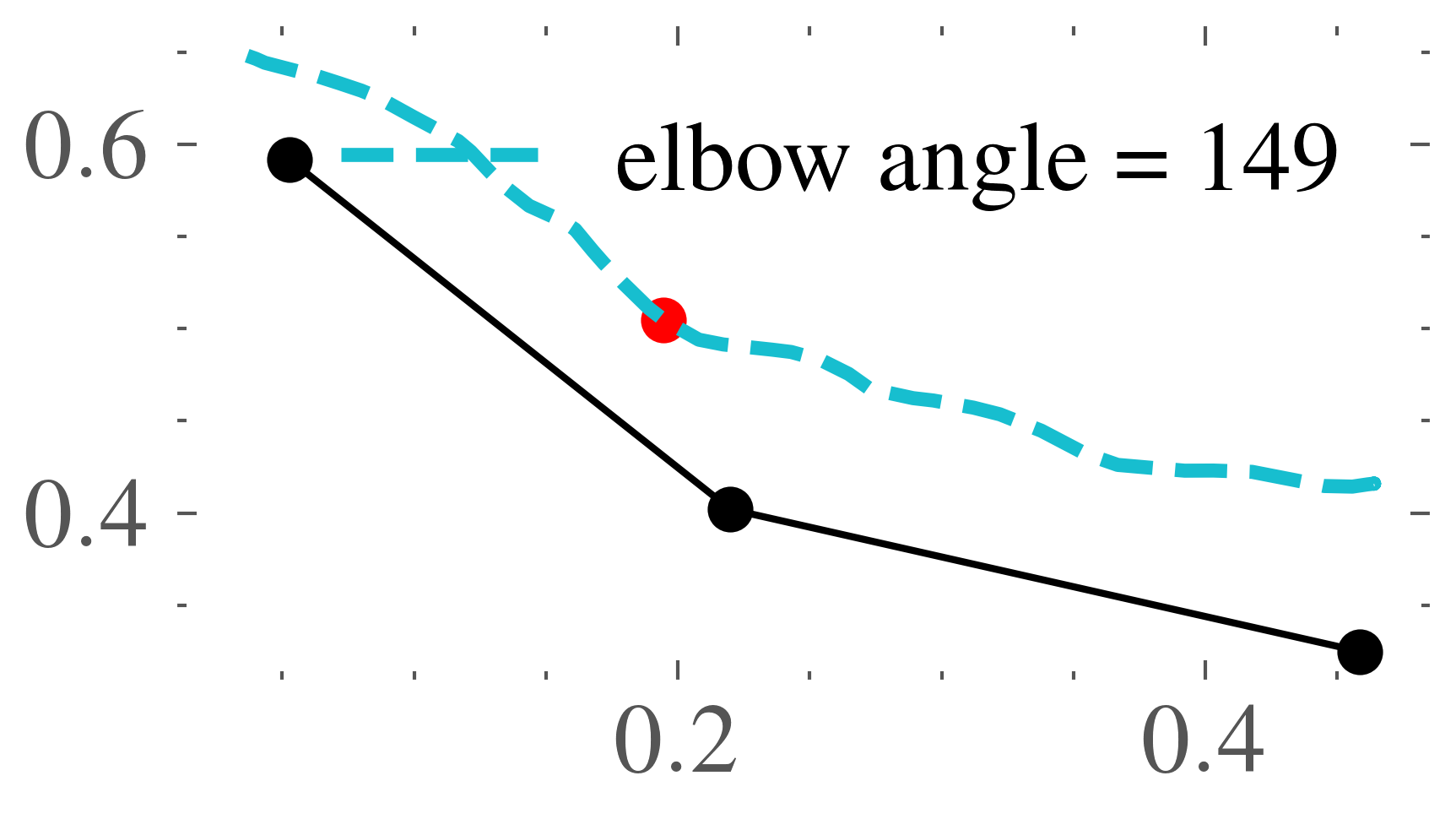}\label{fig:elbow_6}}
    \caption{The projection of the arm postures and the corresponding demonstrated dressing path on the arm plane. The red dot is a transition point corresponding to the middle of the elbow arc on the progress curve.}
    \label{fig:elbow_angle_projection}
\end{figure}

To validate whether the elbow angle affects the dressing strategies, we project the dressing paths onto the arm plane together with the arm postures as shown in Fig.~\ref{fig:elbow_angle_projection}. In the figure, each subfigure corresponds to a 2D projection of the posture and dressing path in Fig.~\ref{fig:elbow_angle_n_dressing_strategies}. The red dot on each subfigure is a transition point corresponding to the middle of the elbow arc in Fig.~\ref{fig:progress_curve} on the progress curve.

In addition, we provide a quantitative result summarized in Tab. \ref{tab:elbow_angle_distance}. We calculate the distance between the transition point and the elbow. When the transition point is on the inner arm plane, the distance is positive which indicates that the dressing employs an inner strategy. When the transition point is on the outer arm plane, we set the distance negative which indicates that the dressing employs an outer strategy (see Sect. \ref{sec:optimal_stretch} and Fig. \ref{fig:inner_outer_path} for the definition of inner/outer strategy).  

\begin{table}[!t]
    \centering
    \caption{Elbow angle and corresponding distance between the transition point and the elbow on the arm plane.}
    \begin{tabular}{p{1.75cm}|c|c|c|c|c|c}
    \hline
        elbow angle (\textdegree) & 80.47 & 81.65 & 87.49 & 106.07 & 143.46 & 149.30 \\
        \hline
        distance (m) & -0.099 & -0.11 & -0.10 & -0.07 & 0.107 & 0.105 \\
        \hline
    \end{tabular}
    \label{tab:elbow_angle_distance}
\end{table}

We can observe both from Fig.~\ref{fig:elbow_angle_projection} and Tab. \ref{tab:elbow_angle_distance} that the outer dressing strategy is employed when the elbow angle is small, which justifies our analysis in Sect.~\ref{sec:elbow_angle}. 

\subsection{Evaluation of the posture estimation scheme}\label{sec:eval_posture_est}
As the dressing path is highly dependent on the arm posture, posture estimation is crucial to the success of dressing. In this section, we provide a quantitative evaluation of the posture estimation scheme. We consider
\begin{enumerate}
    \item the shoulder is static during dressing,
    \item the initial arm posture (in the dressing robot frame $\mathcal{O}^D$) is known. 
\end{enumerate}

The first assumption is a common assumption in assistive dressing literature \cite{zhang2019probabilistic,pignat2017learning}. The initial arm posture in our case is estimated by setting the dressing arm $\mathcal{D}$ in the gravity compensation mode, then putting the end-effector close to the hand, elbow, and shoulder, and using the reading from the end-effector positions as the initial hand, elbow, and shoulder positions. 
In assistive dressing, the starting posture is easy to obtain. It is often estimated via a (depth) camera in the previous literature \cite{gao2016iterative,pignat2017learning,chance2017quantitative,zhang2019probabilistic}.  

With the above assumptions and real-time hand position recovery from the interactive robot (see. Sect.~\ref{sec:human_posture_est}), the only estimated quantity in the full posture description $\vP = \{\vp_{s}, \vp_{e}, \vp_{h}\}$ is the elbow position $\vp_{e}$.

Thus in the evaluation, for a series of arm postures during movements $\{\vP_0, \vP_1,\cdots, \vP_T\}$, we use the maximum error of the elbow as the performance indicator of our posture estimation scheme:
\begin{equation*}
    \text{error}_{\max} = \max_{i = 1}^T{|\vp_{e}(i) - \hat{\vp}_{e}(i)|}  
\end{equation*}
where $\hat{\vp}_{e}(i)$ is the estimated elbow position at the $i^{\text{th}}$ instance.

One tuneable parameter that affects the estimation performance is the diagonal weight matrix $\vQ$ in \eqref{eq:optim_ik_arm_posture}. We compute $\vQ$ using the movement data $\{\vP_0, \vP_1,\cdots, \vP_T\}$. The data is first transformed to angle representation $\{\vq_0, \vq_1,\cdots, \vq_T\}$ (refer to Sect.~\ref{sec:human_posture_est} and Fig.~\ref{fig:arm_posture_est} for the definition of angle representation and $4$ angles $\vq = [\alpha~\beta~\phi~\gamma]^T$ associated). We then compute the difference between adjacent angles and yield:
\begin{equation}
    \{\delta\vq_1, \delta\vq_2, \cdots, \delta\vq_T\},
\end{equation}
where 
\begin{equation}
    \delta \vq_i = [\delta \alpha_i~\delta \beta_i~\delta \phi_i~\delta \gamma_i]^T = \vq_i - \vq_{i-1}.
\end{equation}

The weighting diagonal matrix $\vQ$ in \eqref{eq:optim_ik_arm_posture} is calculated as:
\begin{equation*}
    \vQ = \text{diag}\{\frac{1}{\sum_{i = 1}^T \delta \alpha_i^2},
    \frac{1}{\sum_{i = 1}^T \delta \beta^2},
    \frac{1}{\sum_{i = 1}^T \delta \phi_i^2},
    \frac{1}{\sum_{i = 1}^T \delta \gamma_i^2}\}
\end{equation*}
We conducted experiments to collect arm movement data with an XSens motion capture suit. During the data collection phase, one person is asked to stretch the arm of the other. The posture data was collected using the XSense suit and regarded as ground truth. Using one stretch data, we calculate the weight matrix as (This is the $\vQ$ value used for the interactive dressing experiments in Sect. \ref{sec:interactive_dressing}):
\begin{equation}
    \vQ = \text{diag}\{207.89,~654.28,~99.89,~184.65\}.
\end{equation}

For the evaluation, we collected four new stretch data from different starting poses using the XSense suit. The maximum and average errors with regards to the ground truth in all $4$ cases are presented in Tab. \ref{tab:max_error_elbow_est}. In the same table, we also include estimation errors from \cite{chance2018elbows} for comparison.

In \cite{chance2018elbows}, the authors employed a recurrent neural network (RNN) that trained on posture data during dressing for elbow position estimation under occlusion. 
In Tab. \ref{tab:max_error_elbow_est}, we present estimation errors in \cite{chance2018elbows} which trains on position data only and additionally with engineered features. From the table, we observe that our method performs better than RNN trained with position data only and is comparable to RNN trained on position and feature data.

\begin{table}[t]
    \centering
    \caption{Errors in elbow estimation using our scheme compared with results in the literature.}
    \begin{tabular}{p{2cm}|p{0.4cm}|p{0.4cm}|p{0.4cm}|p{0.4cm}|p{0.8cm}|p{1cm}}
    & \multicolumn{4}{c|}{Our method} & \multicolumn{2}{c}{RNN \cite{chance2018elbows}} \\
     \hline
      & case 1 & case 2 & case 3 & case 4 & position only & with features\\
     \hline
     $\text{error}_{\max}$ (mm) & 28 & 26 & 32 & 28 & 44 & 25\\
     \hline
     $\text{error}_\text{average}$ (mm) & 17 & 14 & 23 & 15 & 41 & 24\\
     \hline
    \end{tabular}
    \label{tab:max_error_elbow_est}
\end{table}

\begin{figure}[b]
    \centering
    \subfloat[case 1]{\includegraphics[width=0.24\textwidth]{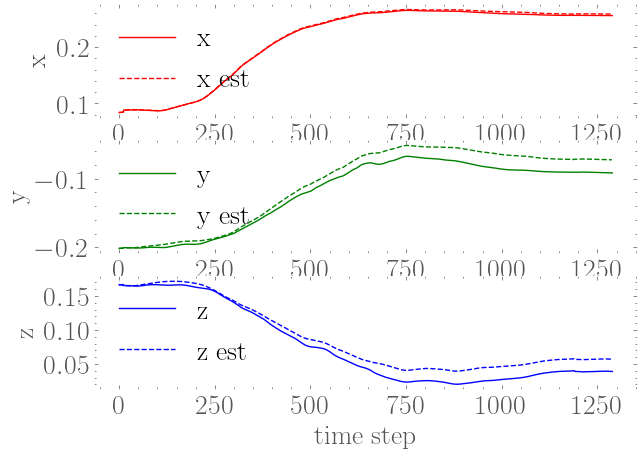}\label{fig:stretch_case_1}}
    \subfloat[case 2]{\includegraphics[width=0.24\textwidth]{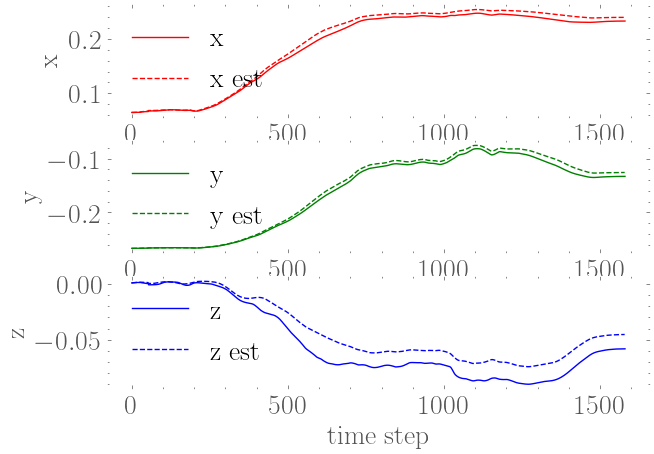}\label{fig:stretch_case_2}} \\
    \subfloat[case 3]{\includegraphics[width=0.24\textwidth]{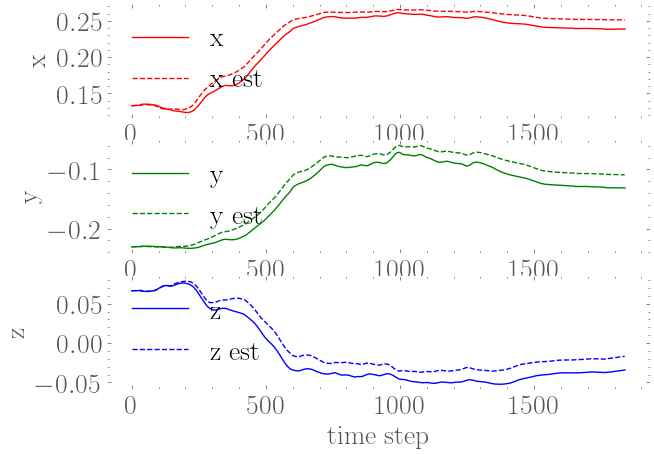}\label{fig:stretch_case_3}} 
    \subfloat[case 4]{\includegraphics[width=0.24\textwidth]{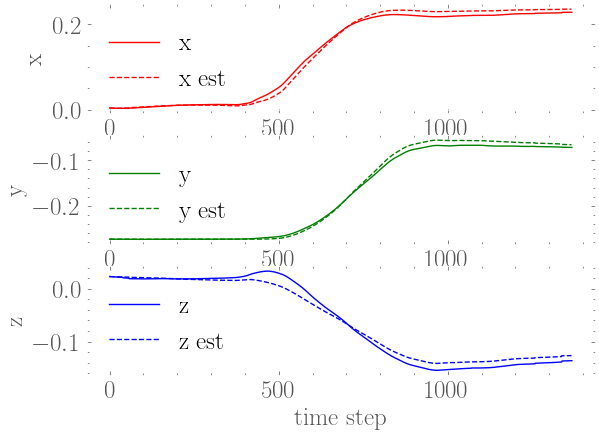}\label{fig:stretch_case_4}}
    \caption{The ground truth and the estimated elbow position ($x,y,z$) in solid and dash line respectively in test cases.}
    \label{fig:stretch_case}
\end{figure}

Figure \ref{fig:stretch_case} shows the ground truth and the estimated elbow position ($x,y,z$) in solid and dash line respectively in all $4$ cases.

We observe that generally, the deviation from ground truth is larger as the time step increases, this is probably due to the recursive nature of the estimation scheme. The error may accumulate with time. Despite this, the maximum error is around $3$ centimeters which is much smaller than the size of the armscye. 
The results are comparable with the results from \cite{chance2018elbows}. 

The arm posture estimation is an adapted solution to allow real-time tracking of human postures during dressing with our proposed bimanual scheme. We are not expecting it to outperform the SOTA but a compatible performance indicates that the framework individually is able to work properly. Later in Sect. \ref{sec:interactive_dressing} and also the video accompanying this paper we will show that the integrated framework can dress different humans and clothes, which in return validates our choice of such an estimation scheme.     

\subsection{Interactive dressing}\label{sec:interactive_dressing}
In this section, we show the experimental results of our interactive dressing scheme. We present experiments with three types of clothes: a short and rigid sleeve shirt, a long and rigid sleeve shirt, and a long and soft sleeve shirt.   

\begin{figure*}[t]
    \centering
    \includegraphics[width=0.9\textwidth]{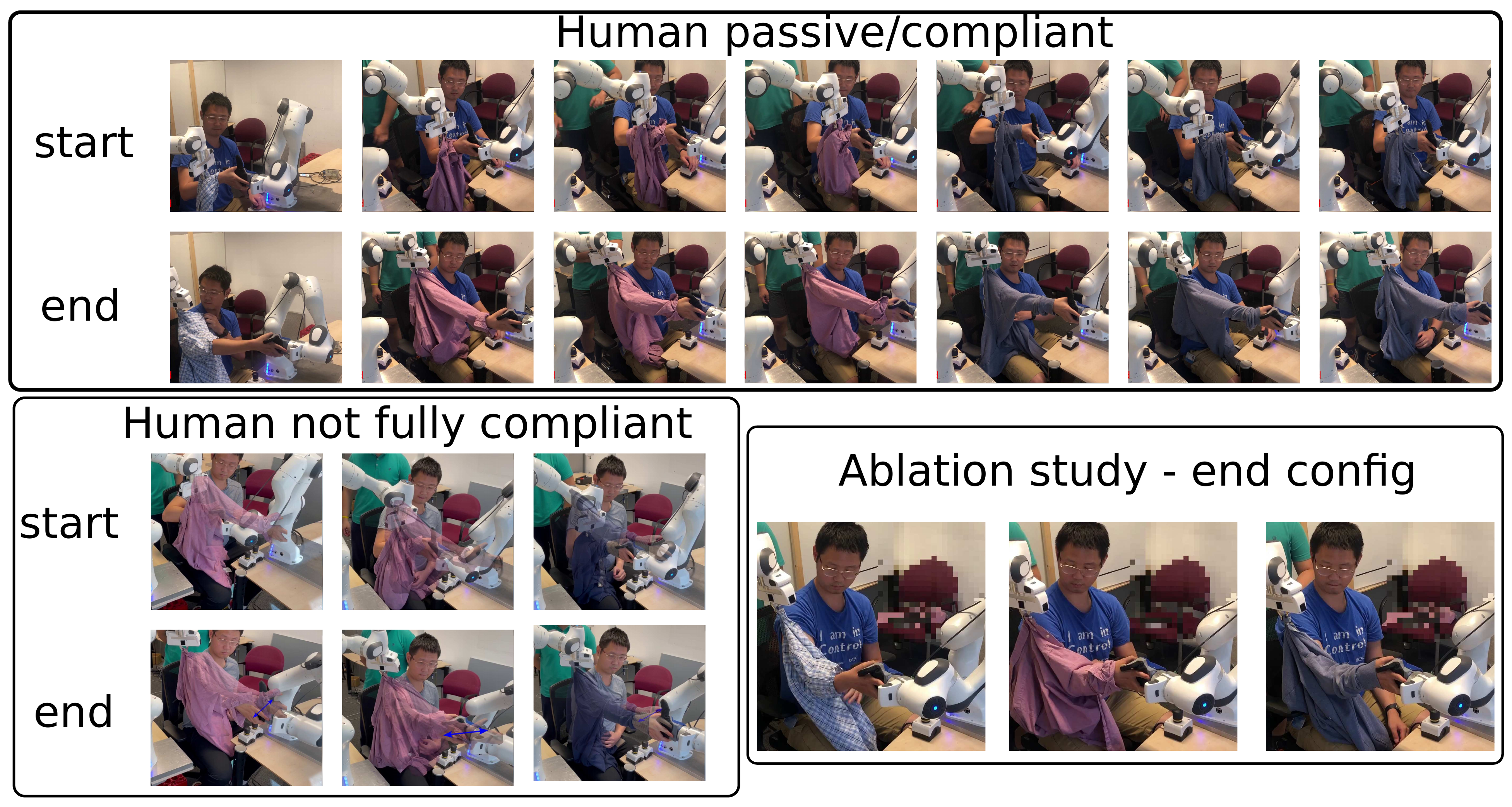}
     \caption{Starting and/or end configurations for $3$ different experiments: human passive/compliant, human not fully compliant, and the ablation study where the human remains static. For experiments where the human is not fully compliant, we also display with high transparency a nominal end configuration when the human is compliant. The deviation from the nominal end configuration is indicated by a blue arrow.}
    \label{fig:robot_exp}
\end{figure*}

To show the flexibility of our framework, our demonstrations were collected on the mannequin and the learned dressing policy was executed on the human arm. In demonstrations, we always start around above the hand and with the same orientation\footnote{The orientation can be seen in the start configuration in dressing in Fig. \ref{fig:robot_exp}}. The arm length of the humans and the mannequin is listed in Tab. \ref{tab:mannequin_human_arm}, although having different lengths, the ratio between hand-elbow and elbow-shoulder is similar. Encoding the policy in the dressing coordinate allows it to be adaptive to different arm lengths.

\begin{table}[t]
    \centering
    \caption{Arm lengths of the human and the mannequin}
    \begin{tabular}{c|c|c|c}
    \hline
      & hand - elbow (cm) & elbow - shoulder (cm) & ratio \\
     \hline
     mannequin & 26.4 & 25.3 & 1.04\\
     \hline
     human 1 & 30.5 & 29.6 & 1.03 \\
     \hline
     human 2 & 27.5 & 26.3 & 1.04 \\
     \hline
    \end{tabular}
    \label{tab:mannequin_human_arm}
\end{table}

\begin{figure*}[!thbp]
    \centering
    \sbox{\measurebox}{%
      \begin{minipage}[t]{.26\textwidth}
      \subfloat[Expert demonstrations on different postures]{\includegraphics[width=\textwidth]{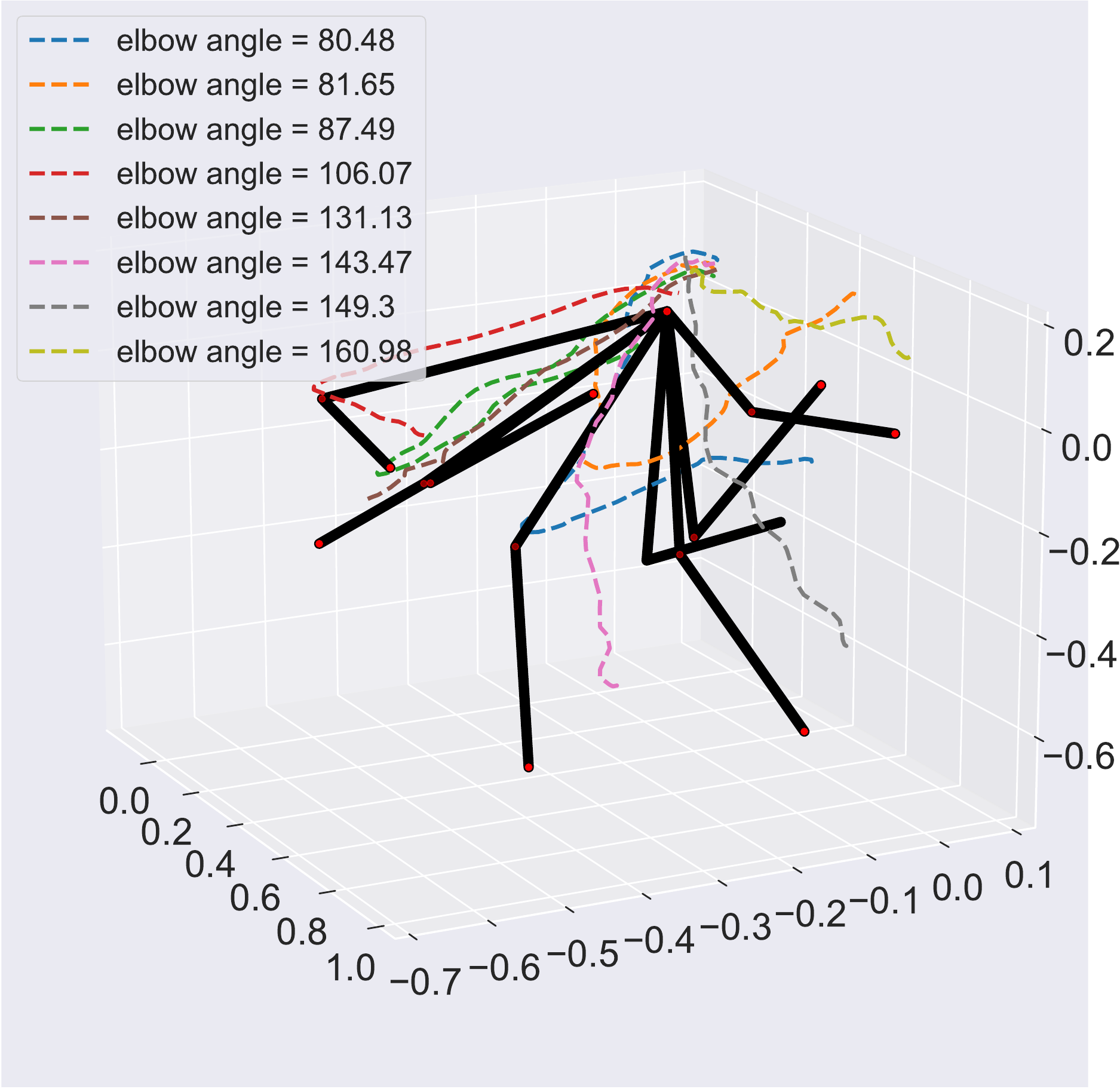}\label{fig:demo_for_training}}
      \end{minipage}}
    \usebox{\measurebox}
    \begin{minipage}[t][\ht\measurebox][s]{.73\textwidth}
    \centering
        \subfloat[Elbow angle = $80.48$]{\includegraphics[width=0.16\textwidth]{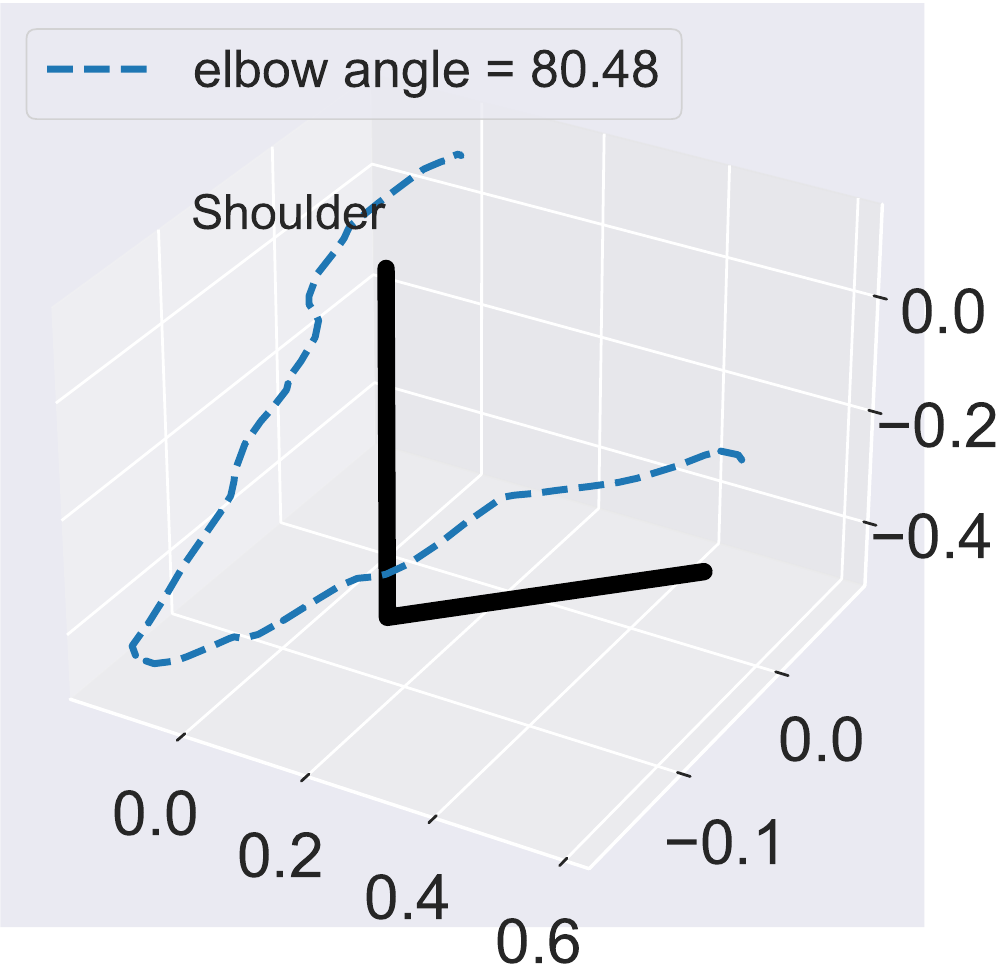}\label{fig:traning_data_1}}
        \hspace{0.3mm}
        \subfloat[Elbow angle = $81.65$]{\includegraphics[width=0.16\textwidth]{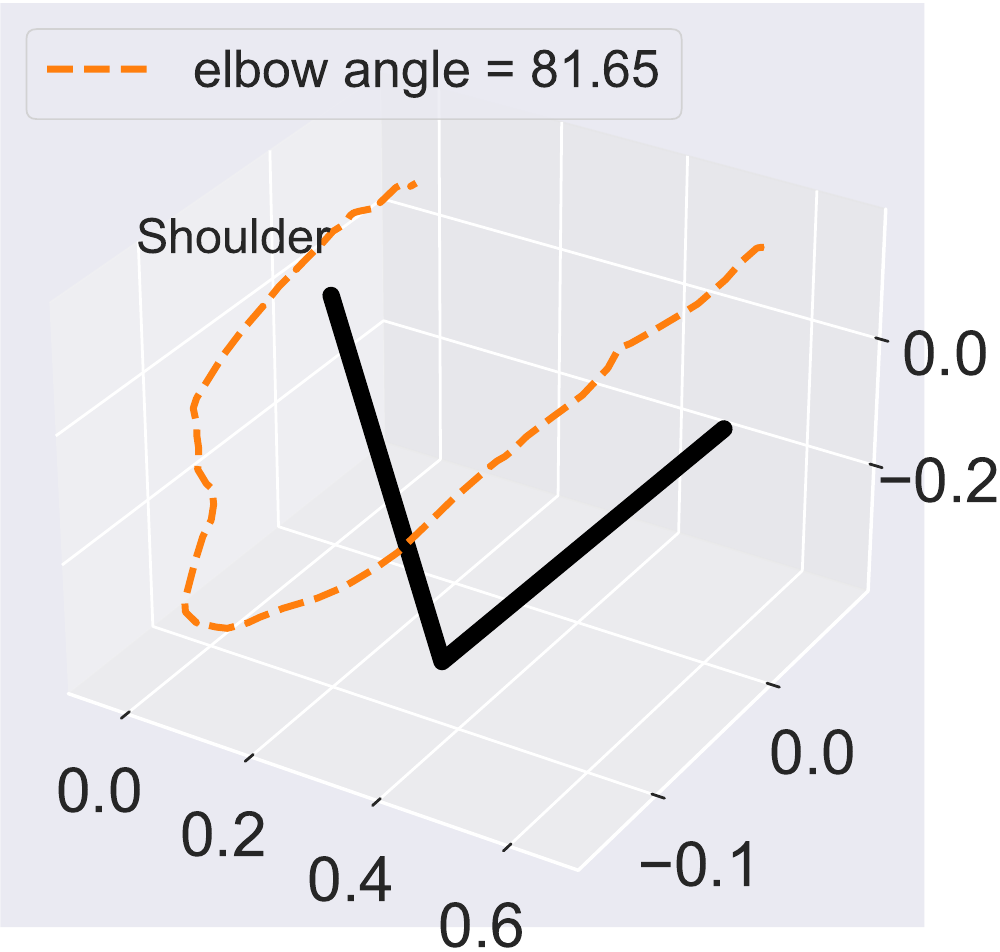}\label{fig:traning_data_2}}
        \hspace{0.3mm}
        \subfloat[Elbow angle = $87.49$]{\includegraphics[width=0.16\textwidth]{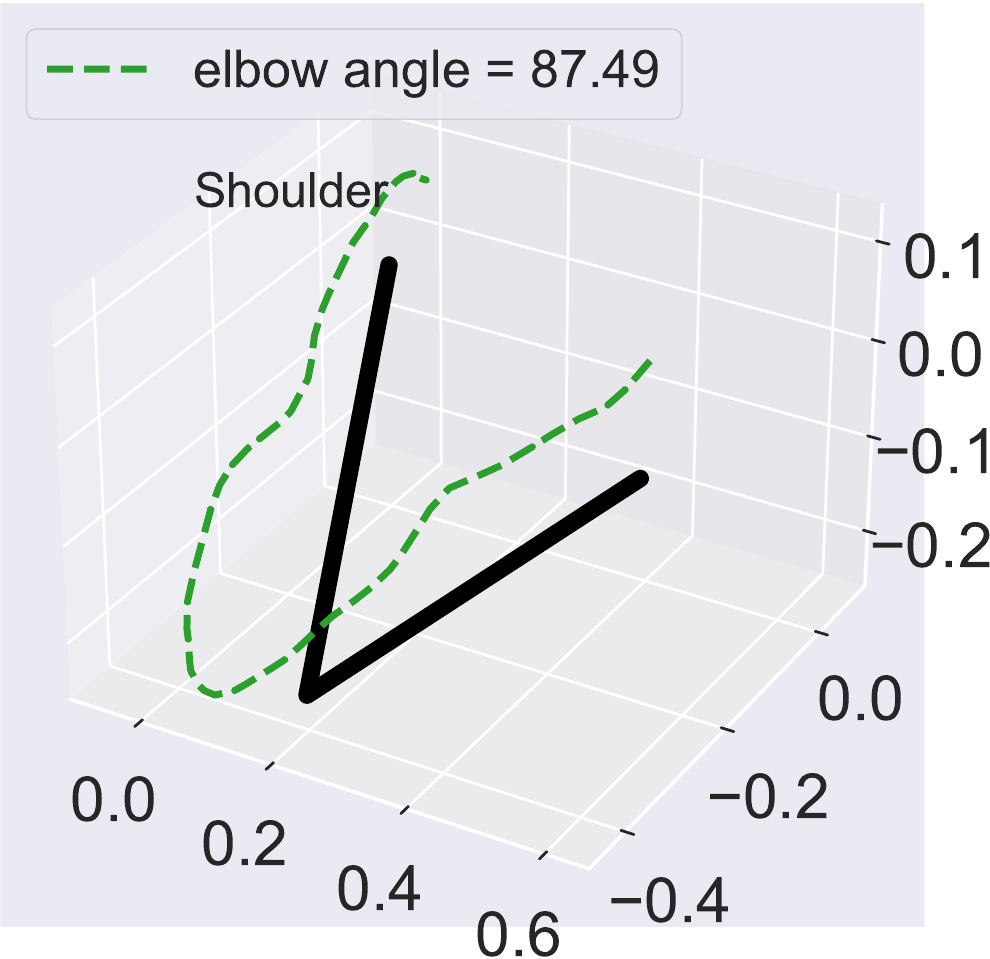}\label{fig:traning_data_3}}
        \hspace{0.3mm}
        \subfloat[Elbow angle = $106.07$]{\includegraphics[width=0.17\textwidth]{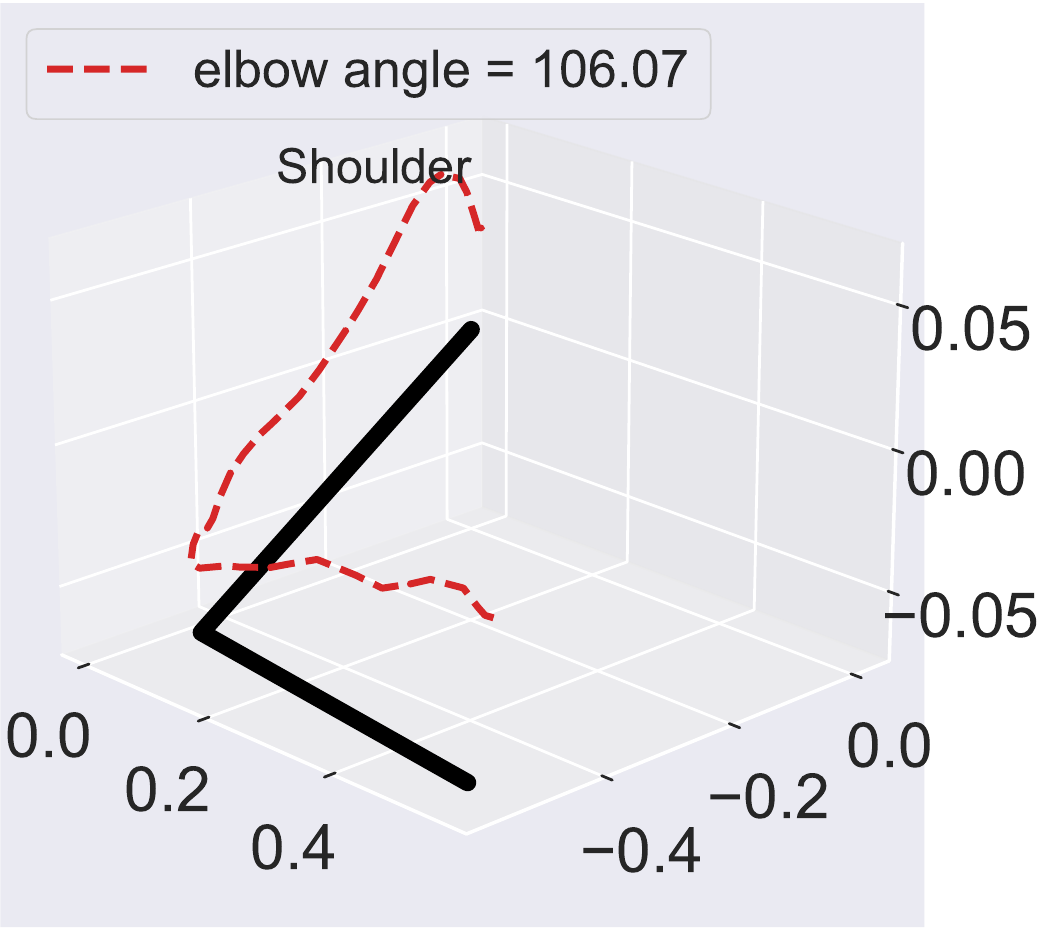}\label{fig:traning_data_4}} \\
        \subfloat[Elbow angle = $131.13$]{\includegraphics[width=0.16\textwidth]{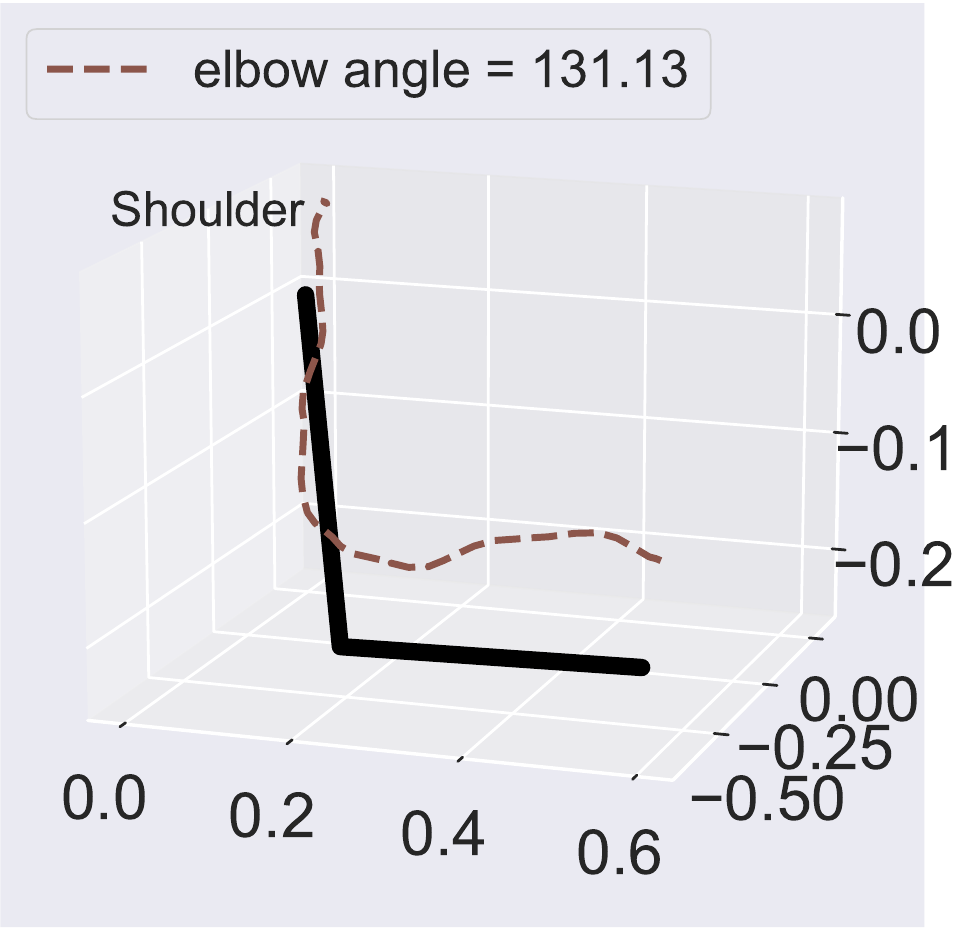}\label{fig:traning_data_5}}
        \hspace{0.3mm}
        \subfloat[Elbow angle = $143.47$]{\includegraphics[width=0.16\textwidth]{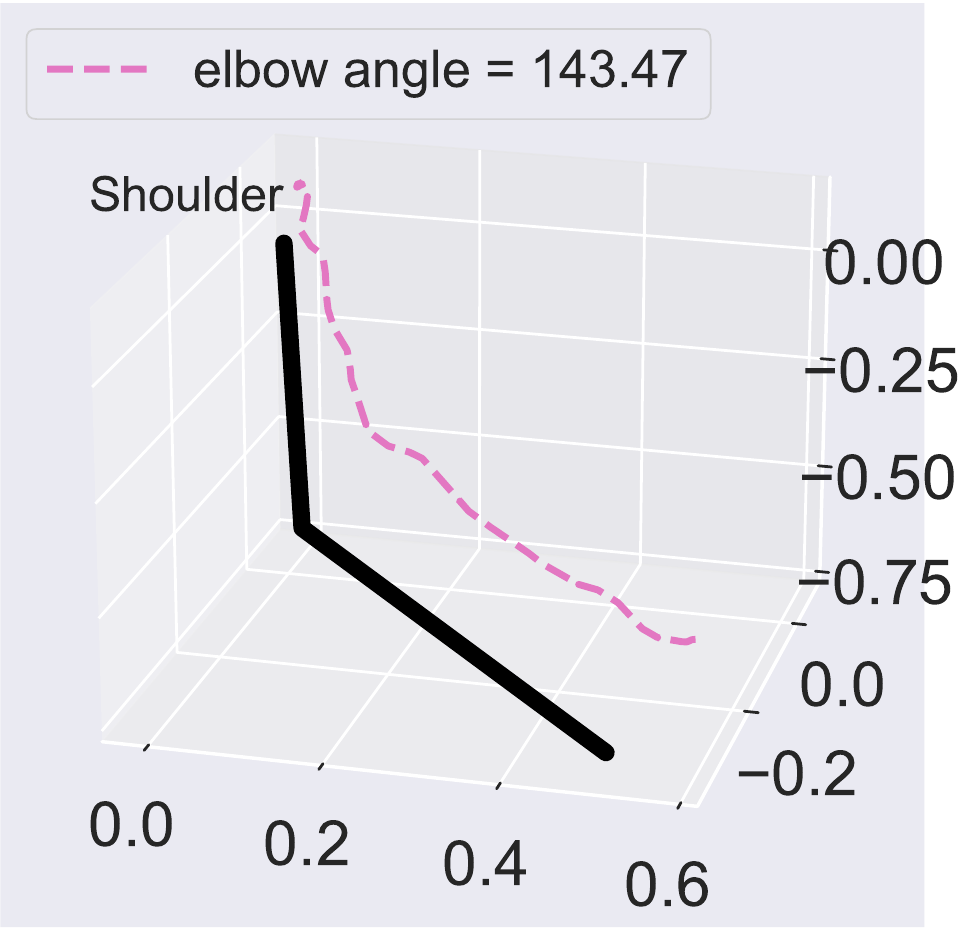}\label{fig:traning_data_6}}
        \hspace{0.3mm}
        \subfloat[Elbow angle = $149.3$]{\includegraphics[width=0.16\textwidth]{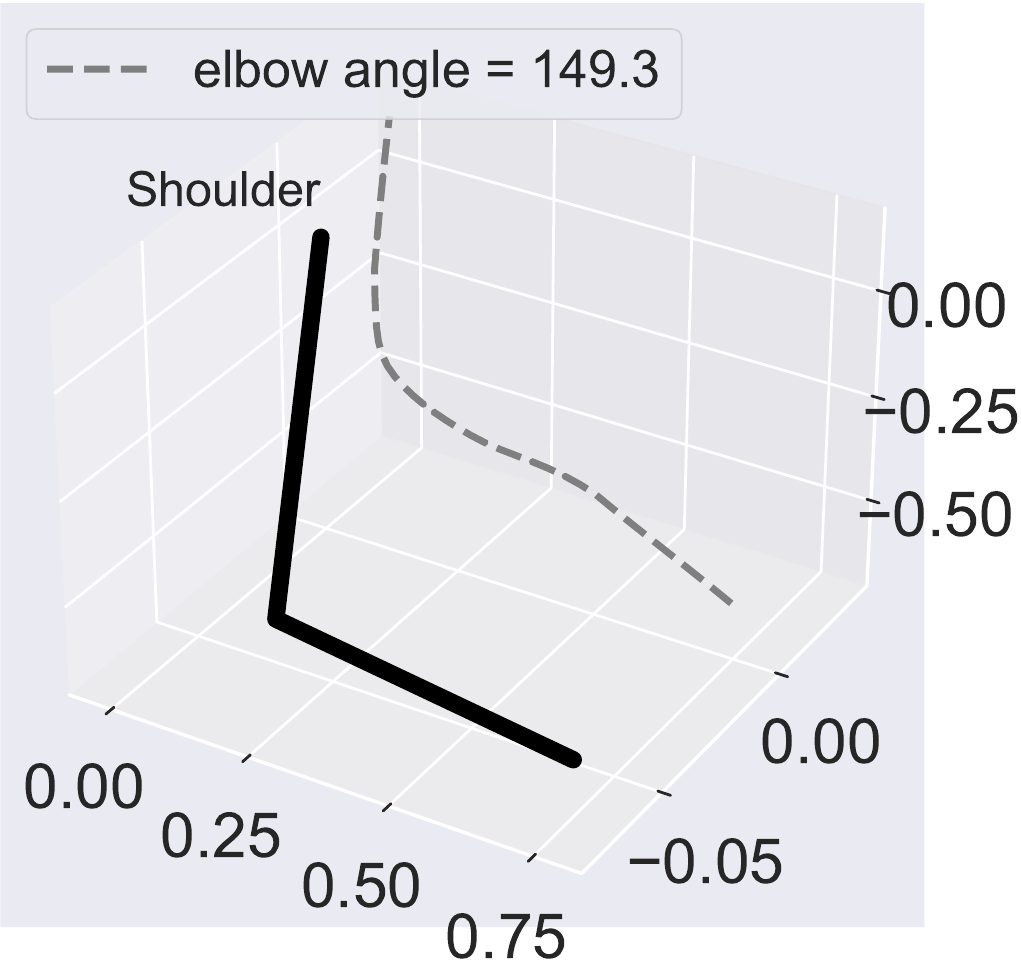}\label{fig:traning_data_7}}
        \hspace{0.3mm}
        \subfloat[Elbow angle = $160.98$]{\includegraphics[width=0.16\textwidth]{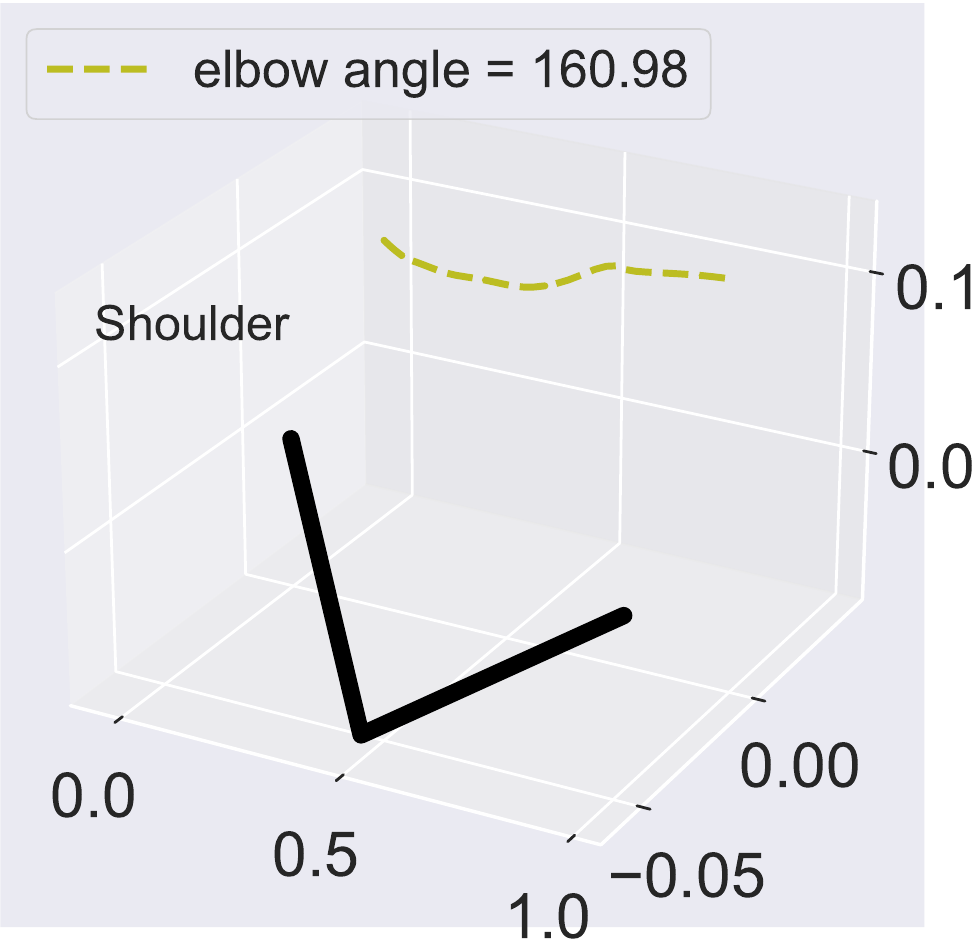}\label{fig:traning_data_8}}
    \end{minipage}
    \vspace{6mm}
    \caption{The expert demonstrations on different postures used for training dressing policy in (a), and individual trajectory in (b) -- (i).}
    \label{fig:training_data}
\end{figure*}

Fig.~\ref{fig:training_data} shows the demonstration and transformed data in the dressing coordinate. For the angle around the arm $\theta$ and the distance to the arm $l$, we present the changes during dressing $\delta \theta$ and $\delta l$ instead of the absolute values. The difference is computed with regard to the starting position which is around above the hand. For postures with smaller elbow angles, we can observe a change in the angles around the arm $\theta$ since it needs to move to the back of the arm to execute an outer strategy. 

\begin{figure}[!t]
    \centering
    \includegraphics[width=0.4\textwidth]{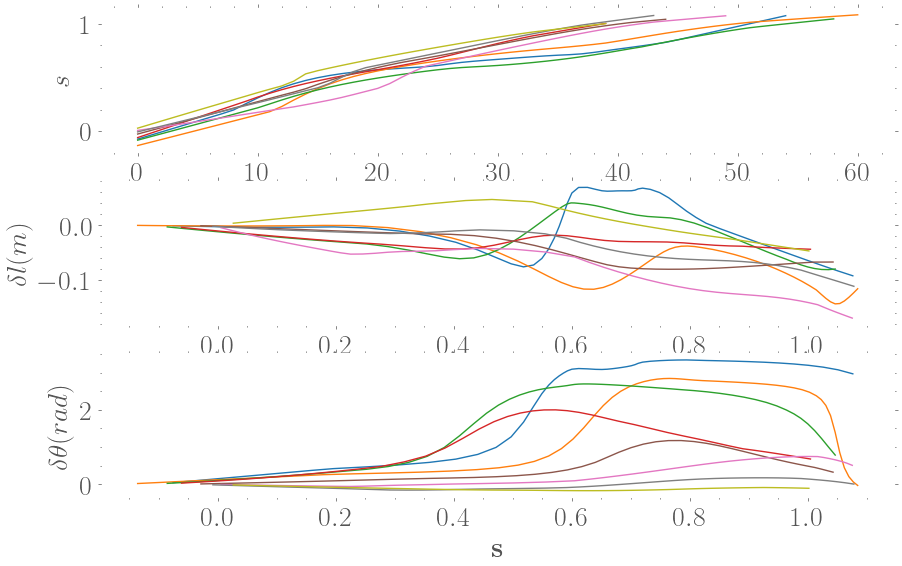}
    \caption{The expert demonstrations for training the dressing policy in Fig. \ref{fig:demo_for_training} transformed to dressing coordinate.}
    \label{fig:demo_in_dress_coord}
\end{figure}
From Fig.~\ref{fig:demo_in_dress_coord} we observe that the progress scalar $s$ is a monotonic increasing quantity. Consider a target $s_{\text{target}}$, we model $s$ with a dynamical system with a constant increment $c > 0$:
\begin{equation}\label{eq:s_dy}
    s_{T + 1} = s_T + \min(c,~(s_{\text{target}} - s_T))
\end{equation}

\begin{figure}[t]
    \centering
    \subfloat[The 3D surface plot of GMM for encoding $\delta l$]{\includegraphics[height=3.8cm]{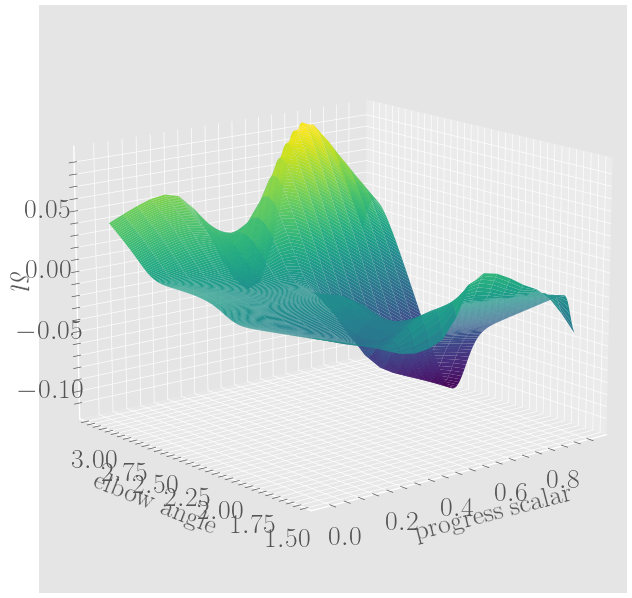}\label{fig:GMM_dl}}
    \hspace{2mm}
    \subfloat[The 3D surface plot of GMM for encoding $\delta \theta$]{\includegraphics[height=3.8cm]{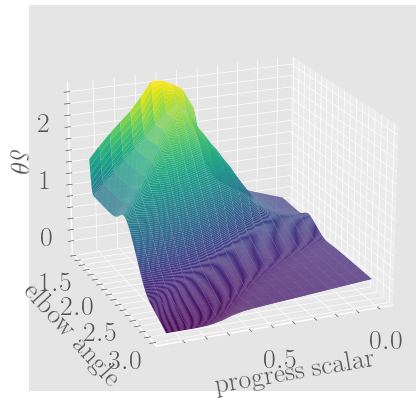}\label{fig:GMM_dtheta}} \\
    \caption{The surface plot for GMM with elbow angles $\psi$ and progress scalar $s$ as inputs and outputs changes in  the distance to the arm $\delta l$ and changes in angles around the arm $\delta \theta$ respectively.}
    \label{fig:GMM_models}
\end{figure}

We respectively trained two GMM for the changes in angles around the arm $\delta \theta$ and changes in the distance to the arm $\delta l$ with inputs progress scalar $s$ and the elbow angle $\phi$\footnote{Please note that other formulations may be also possible, such as to learn the joint distribution of $[s,~\phi,~l,~\theta]$}. We use $8$ Gaussians for both $\delta l$ and $\delta \theta$. The resulting mixture models are presented in 3D surf plots in Fig.~\ref{fig:GMM_models}. We can observe a smaller elbow angle displays larger changes in $\theta$ which is a clear indication for taking the outer strategy. For a large elbow angle, the $\theta$ change is almost zero which resembles an inner strategy. 
Due to the diminishing rigidity, the end effector rotation has a very limited effect during dressing. In the experiment, we consider only the $z$ axis orientation of the end-effector of the dressing robot. The $z$ axis rotation is applied to avoid collision with the arm during dressing. We can use a constant incremental rotation with regard to $s$ or a separate GMM with the same inputs ($[s,~\phi]$) for encoding the difference $dz$ in $z$ axis rotation (The training procedure is the same as $\delta \theta$ and $\delta l$). 

We implement the policy with our optimal stretch and posture estimation scheme. Three types of experiments are conducted:
\begin{enumerate}
    \item Human passively follows the interactive robot
    \item Human is not fully compliant with the guiding force and moves in other directions
    \item Ablation study: human arm remains static
\end{enumerate}

We tested in different starting postures and different clothes as shown in Fig.~\ref{fig:robot_exp}. In the figure, we present only starting and end configurations, the full dressing sequences are recorded in the video accompanying the paper.
We tested the dressing while the human passively follows the guiding force and also moves in other directions (not fully compliant). When the human is not fully compliant, in the figure, we also display with high transparency a nominal end configuration when the human is compliant. The deviation from the nominal end configuration is indicated by a blue arrow. With these results, we prove that humans can move in other directions if desired and the framework is robust against non-compliant behavior and still succeed in dressing. 

In the ablation study, we demonstrate cases where the elbow angle is small, and the cloth tends to get stuck around the elbow, thus reiterating the importance of having a bimanual setup for dressing. However, the result also suggests that if the human starts in these postures and refuses to move along with the guiding force, the dressing could fail.

Lastly, we compare our dressing coordinate encoding with TPGMM that were commonly used to encode dressing behavior from demonstration \cite{hoyos2016incremental, pignat2017learning, zhu2022learning} (while the rest of the framework stays the same), and present respectively the success rates in dressing while the human is compliant for both methods in Tab. \ref{tab:exp_success_rate}. There are three main reasons for failure:
\begin{itemize}
    \item The clothes slip from the gripper,
    \item The trajectory collides with humans,
    \item Robot stops before reaching the shoulder.
\end{itemize} 
As clothes grasping is not the focus of this paper, in Tab. \ref{tab:exp_success_rate}, we show the overall success rate in the fourth column and the success rate excluding gripper slip in the fifth column. Our dressing coordinate approach achieves a higher success rate.

\begin{table}[!thpb]
    \centering
    \caption{Comparison between dressing coordinate-based approach and TPGMM}
    \begin{tabular}{p{1.78cm}|p{1.2cm}|p{0.75cm}|p{0.75cm}|p{1.8cm}}
    \hline
      Methods & number of trials & success trials & success rate & success rate excl. slipping \\
     \hline
     Dressing Coord & 12 & 9 & 75\% & 90\%  \\
     \hline
     TPGMM & 6 & 3 & 50\% & 50\%  \\
     \hline
    \end{tabular}
    \label{tab:exp_success_rate}
\end{table}

The reasons why TPGMM is more prone to fail compared with our dressing coordinate approach are:
\begin{enumerate}
    \item Due to covariance shifts, traditional GMM-based learning cannot guarantee convergence. Although there exist methods that combine dynamical systems with GMM to guarantee convergence \cite{khansari2011learning}, such methods do not work for TPGMM. While using our method to learn in dressing coordinate, there is one progress scalar that directly encodes the progress of the dressing. The scalar is defined as monotonically increasing, thus the trajectory will converge around the shoulder.
    \item TPGMM requires the elbow position to determine the reference frames for calculating the final motion generation model, therefore, the error in the elbow estimation will be directly reflected in the motion generation model. 
\end{enumerate}

\section{Conclusion}\label{sec:conclusion}
Traditional assistive dressing considers a \textit{one-robot-to-one-arm} setup, which 
\begin{itemize}
    \item either assumes static arm posture or requires additional sensors and algorithms for tracking the arm posture,
    \item renders the human arm hanging in the air during the dressing process which is often tiring,
    \item may fail/get stuck at the elbow for some initial arm postures.
\end{itemize}

In the light of the following limitations, and inspired by strategies taken by caregiving experts, we designed a bimanual assistive dressing framework with a \textit{two-robot-to-one-arm} setup that
\begin{itemize}
    \item allows tracking of arm postures without additional sensors,
    \item provides support and guiding force for the arm being dressed,
    \item stretches the human arm while compliant to obtain postures for the ease of dressing. 
\end{itemize}

We approached the design by first analysing the effect of elbow angle in dressing. Based on this effect, we proposed an optimal stretch controller for the interactive robot. As the interactive robot holds the human hand, the arm posture can be estimated in real-time with inverse kinematics. For the dressing robot, we utilize the dependency of the dressing policy on the arm posture and introduced a dressing coordinate defined by the arm posture for easy encoding of the dressing policy from demonstrations.

The adoption of the dressing coordinate greatly enhances the flexibility of the LfD. The trained policy can adapt to different arm lengths as long as the forearm to upper-arm ratio is similar. The framework allows dressing different types of clothes, even long sleeves which is unprecedented in previous research. The dressing remains robust even though human is not fully compliant with the guidance from the interactive robot. 

Since this is the first time such a bimanual setup was employed in dressing a human, there are two major limitations of the proposed framework:
\begin{itemize}
    \item The assumption of the shoulder remains static is not always true, since the interactive robot is stretching the arm, the shoulder is more likely to move forward. A forward-moving shoulder will subsequently increase the length of the arm (as now the hand can reach further when fully stretched). The increased length may render the arm posture estimation scheme unsolvable thus the overall scheme will fail.
    \item The elbow angle is identified as the crucial parameter that affects dressing, however, there is no explicit coordination between the dressing and interactive robot to make sure that when the dressing robot is going through the elbow, the interactive robot already makes the elbow angle sufficiently large to avoid getting stuck.
\end{itemize}
It is difficult to solve the first limitation purely from the algorithm side, one possibility is to combine inverse kinematics with vision-based tracking for more accurate posture estimation. The second limitation is even trickier, to make the coordination explicit, a thorough analysis of dressing on different elbow angles needs to be conducted to determine a boundary condition of the elbow angle, then additionally, an assumption has to be placed on the human side (passive for instance) to make sure that the interactive robot will always succeed in bringing the human arm into favorable postures before dressing arm reach the elbow.

Another possible direction for future work is safety analysis. We did not explicitly analyze the safety of the overall system. However, for assistive robots, it is crucial to ensure human safety during robotic assistance. Inspirations for conducting system-level safety analysis can be drawn from \cite{mcdermid2019towards}.

Despite the above-mentioned limitations, our proposal is the first that considers interactive dressing assistance with a bimanual setup. The setup is inspired by caregiving experts conducting the task, which makes the scheme receptive to humans. It represents a paradigm shift in thinking of the dressing task from a \textit{one-robot-to-one-arm} setup to a \textit{two-robot-to-one-arm} setup.
\section*{Acknowledgments}
This work was supported by Honda Research Institute Europe GmbH as part of the project ``Learning Physical Human-Robot Cooperation Tasks'' and has been partially funded by the European Research Council Starting Grant TERI ``Teaching Robots Interactively'', project reference \#804907.

\section*{Appendix}
In the Appendix, we provide the analytical solution to \eqref{eq:optim_ik_arm_posture} in Sect.~\ref{sec:human_posture_est}:
\begin{equation*}
    \begin{aligned}
        & \min_{\Delta \vq} \vL = \Delta \vq^T \vQ \Delta \vq \\
        & \text{s.t. }\Delta \vp_h = \vJ\Delta \vq
    \end{aligned}
\end{equation*}
Since $\vJ \in \mathbb{R}^{3 \times 4}$, $\Delta \vp_h = \vJ \Delta \vq$ has infinite number of solutions:
\begin{equation}\label{eq:sol_Delta_q}
    \Delta \vq = \vJ^+ \Delta \vp_h + \lambda \vmu = \Delta \vq_h + \lambda \vmu
\end{equation}
where $\vJ^+$ is the Moore–Penrose inverse of the $\vJ$ and $\vmu$ is the vector in the null space of $\vJ$ (e.g., $\vJ \vmu =  \vnull$), and $\lambda \in \mathbb{R}$ is a gain. Taking (\ref{eq:sol_Delta_q}) into $\vL$:
\begin{equation}\label{eq:sol_Delta_q}
    \begin{aligned}
         \vL & = (\Delta \vq_h + \lambda \vmu)^T \vQ ( \Delta \vq_h + \lambda \vmu) \\
             & = (\vmu^T \vQ \vmu) \lambda^2 + 2 \vmu^T \vQ \Delta \vq_h \lambda + \Delta \vq_h^T \vQ \Delta \vq_h
    \end{aligned}
\end{equation}
which is quadratic in $\lambda$.

The original optimization problem can be transformed into:
\begin{equation}
    \min_{\lambda} \vL = (\vmu^T \vQ \vmu) \lambda^2 + 2 \vmu^T \vQ \Delta \vq_h \lambda + \Delta \vq_h^T \vQ \Delta \vq_h \\
\end{equation}
Let us denote $\vmu^T \vQ \vmu = a$, $\vmu^T \vQ \Delta \vq_h = b$ and $\Delta \vq_h^T \vQ \Delta \vq_h = c$:
\begin{equation}
    \begin{aligned}
         \vL & = a \lambda^2 + 2 b \lambda + c \\
             & = a(\lambda^2 + 2\frac{b}{a}\lambda + c) \\
             & = a((\lambda + \frac{b}{a})^2 + c - \frac{b^2}{a^2}) 
    \end{aligned}
\end{equation}
The solution that minimize $\vL$ is: 
\begin{equation}\label{eq:sol_lambda}
    \begin{aligned}
         \lambda & = -\frac{b}{a} = - {\vmu^T \vQ \Delta \vq_h}/{\vmu^T \vQ \vmu}
    \end{aligned}
\end{equation}
Taking (\ref{eq:sol_lambda}) into (\ref{eq:sol_Delta_q}) we can obtain the optimal solution:
\begin{equation}
    \Delta \vq^* = \vJ^+ \Delta \vp_h - ({\vmu^T \vQ \Delta \vq_h}/{\vmu^T \vQ \vmu}) \vmu
\end{equation}

The first term ensures that the constraint is respected; on the other hand, the second term generates a null-space transition that minimizes the total weighted joint displacement. 

\bibliography{ref.bib}
\bibliographystyle{IEEEtran}

\begin{IEEEbiography}
[{\includegraphics[width=1in,height=1.25in,clip,keepaspectratio]{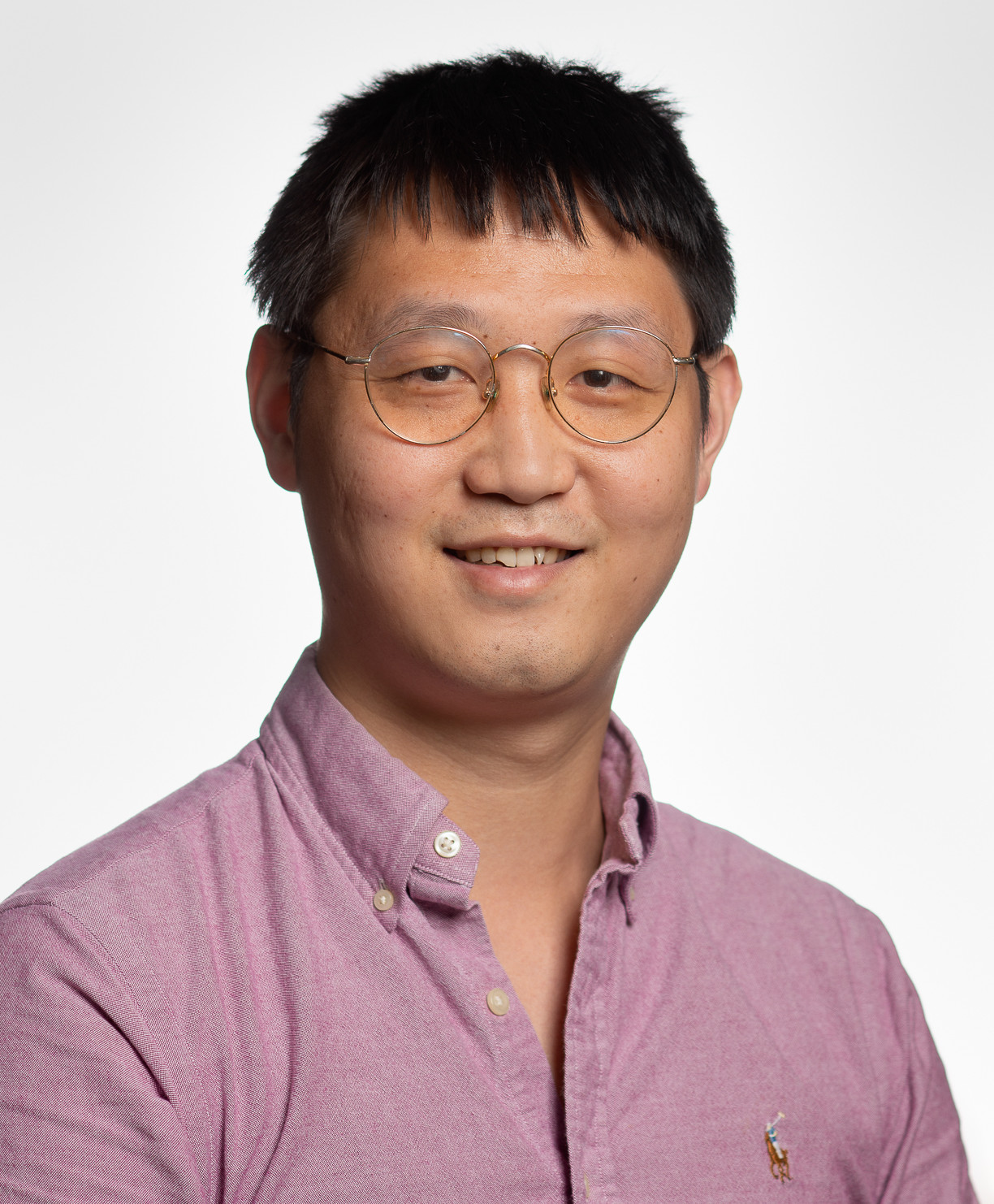}}]
{Jihong Zhu} obtained his Ph.D. in robotics at the University of Montpellier and conducted his research at LIRMM, France. He was a postdoc jointly at Cognitive Robotics, TU Delft, Netherlands, and Honda Research Institute Europe. He is currently an Assistant Professor in robotics at the School of Physics, Engineering, and Technology and an affiliated researcher at the Insitute for Safe Automony, University of York, the UK. He is also a visiting researcher at Cognitive Robotics, TU Delft. He serves as an Associate Editor of the \textsc{IEEE Robotics and Automation Letters}.
\end{IEEEbiography}

\begin{IEEEbiography}
[{\includegraphics[width=1in,height=1.25in,clip,keepaspectratio]{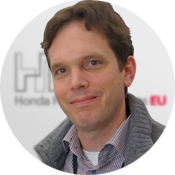}}]
{Michael Gienger} received the diploma degree in Mechanical Engineering from the Technical University of Munich, Germany, in 1998. From 1998 to 2003, he was research assistant at the Institute of Applied Mechanics of the TUM, and received his PhD degree with a dissertation on ``Design and Realization of a Biped Walking Robot''. After this, Michael Gienger joined the Honda Research Institute Europe in Germany in 2003. Currently he works as a Chief Scientist and Competence Group Leader in the field of robotics. His research interests include mechatronics, robotics, whole-body control, imitation learning and human-robot interaction.
\end{IEEEbiography}

\begin{IEEEbiography}
[{\includegraphics[width=1in,height=1.25in,clip,keepaspectratio]{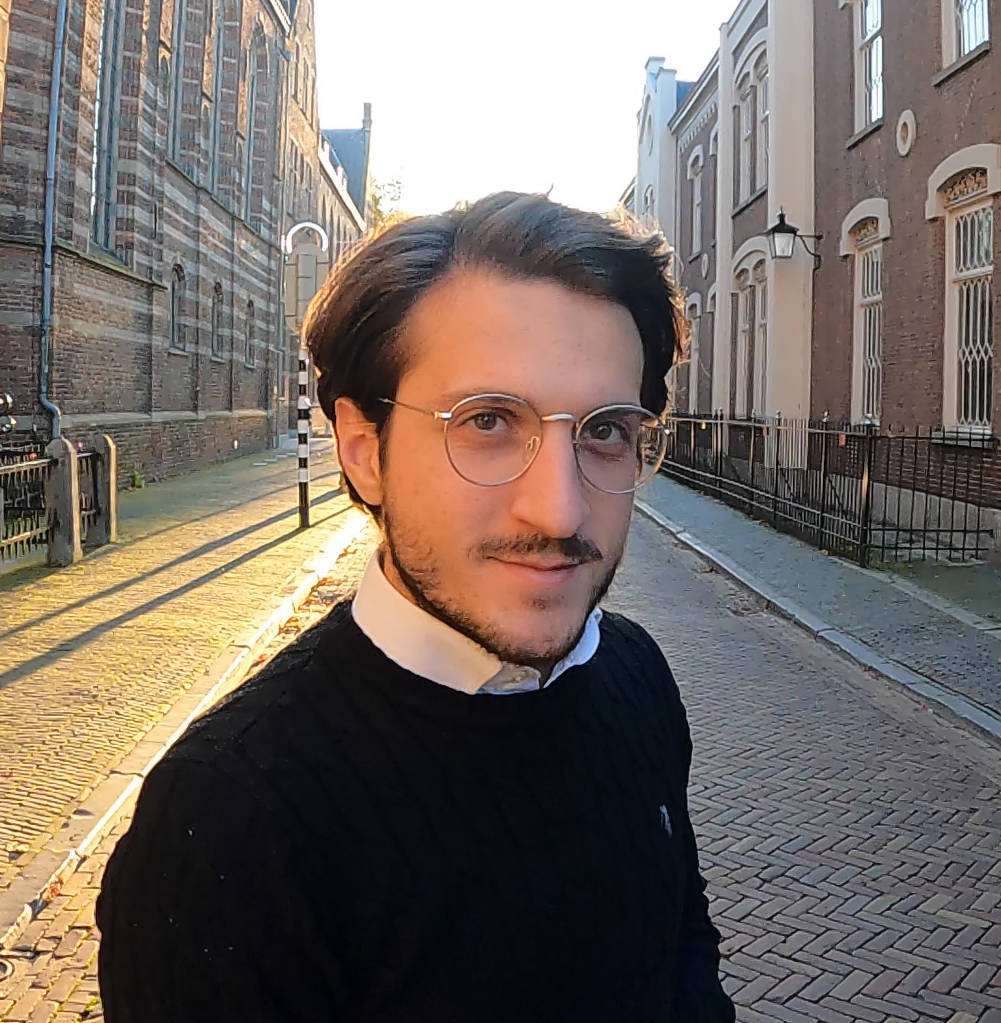}}]
{Giovanni Franzese} is a PhD Student in the Department of Cognitive Robotics at TU Delft, Netherlands since 2019. He received a BSc degree (2016) in Mechanical Engineering and an MSc degree (2018) in Mechatronics and Robotics at Politecnico di Milano, Italy. Since 2022, he has been an ELLIS member for artificial intelligence. His research focuses on Interactive Imitation Learning applied to robot manipulation.  
\end{IEEEbiography}

\begin{IEEEbiography}
[{\includegraphics[width=1in,height=1.25in,clip,keepaspectratio]{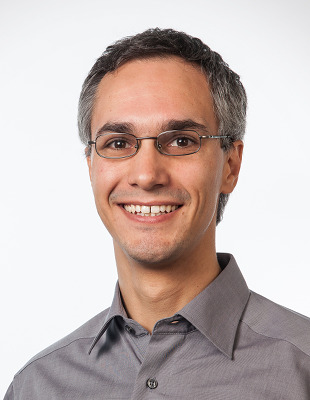}}]
{Jens Kober} received the Ph.D. degree in engineering from TU Darmstadt, Darmstadt, Germany, in 2012.

He is currently an Associate Professor with TU Delft, Delft, The Netherlands. He was a Postdoctoral Scholar jointly with CoR-Lab, Bielefeld University, Bielefeld, Germany, and with Honda Research Institute Europe, Offenbach, Germany.

Dr. Kober was a recipient of the annually awarded Georges Giralt PhD Award for the best
PhD thesis in robotics in Europe, the 2018 IEEE RAS Early Academic Career Award, the 2022 RSS Early Career Award, and was a recipient of an ERC Starting grant.
\end{IEEEbiography}

\vfill

\end{document}